%% file: main.tex
\documentclass[sigconf]{acmart}
\usepackage{amsmath}
\usepackage{amsfonts}
\usepackage{xcolor}
\usepackage{xspace}
\usepackage{graphicx}
\usepackage{subcaption}
\usepackage{multirow}

\newcommand{\ours}{\textsc{NUCLEUS}\xspace}

\AtBeginDocument{%
  }

\def\deg{°}

\copyrightyear{2026}
\acmYear{2026}
\setcopyright{cc}
\setcctype{by}
\acmConference[KDD '26]{Proceedings of the 32nd ACM SIGKDD Conference on Knowledge Discovery and Data Mining V.2}{August 09--13, 2026}{Jeju Island, Republic of Korea}
\acmBooktitle{Proceedings of the 32nd ACM SIGKDD Conference on Knowledge Discovery and Data Mining V.2 (KDD '26), August 09--13, 2026, Jeju Island, Republic of Korea}
\acmDOI{10.1145/3770855.3819026}
\acmISBN{979-8-4007-2259-2/2026/08}

\begin{document}

\title{\ours-MoE: Unified Model of Pool Boiling for Liquid Cooling}

\author{Arthur Feeney}
\orcid{0009-0006-0011-3461}
\affiliation{
    \department{Department of Electrical Engineering and Computer Science}
    \institution{University of California, Irvine}
    \city{Irvine}
    \state{CA}
    \country{USA}
}

\author{Xianwei Zou}
\orcid{0009-0007-7409-3237}
\affiliation{
    \department{Department of Electrical Engineering and Computer Science}
    \institution{University of California, Irvine}
    \city{Irvine}
    \state{CA}
    \country{USA}
}

\author{Sheikh Md Shakeel Hassan}
\orcid{0009-0003-1819-9361}
\affiliation{
    \department{Department of Electrical Engineering and Computer Science}
    \institution{University of California, Irvine}
    \city{Irvine}
    \state{CA}
    \country{USA}
}

\author{Siddhartha Rachabathuni}
\orcid{0009-0009-8541-8361}
\affiliation{
    \department{Department of Electrical Engineering and Computer Science}
    \institution{University of California, Irvine}
    \city{Irvine}
    \state{CA}
    \country{USA}
}

\author{Aparna Chandramowlishwaran}
\orcid{0000-0003-0840-4192}
\affiliation{
    \department{Department of Electrical Engineering and Computer Science}
    \institution{University of California, Irvine}
    \city{Irvine}
    \state{CA}
    \country{USA}
}

\renewcommand{\shortauthors}{}

\begin{abstract}
Two-phase boiling enables heat transfer rates an order of magnitude higher than single-phase cooling, but it remains difficult to model due to the strong coupling between phase change, turbulence, and transport, as well as extreme sensitivity to fluid properties and thermodynamic conditions. 
Existing learning-based surrogates are either condition- or fluid-specific, limiting generalization and requiring separate models. 
We present \ours, a mixture-of-experts model for pool boiling that replaces collections of specialized surrogates with a single architecture. 
\ours combines neighborhood attention, signed distance field reinitialization for interface consistency, and expert routing that exhibits emergent specialization across distinct boiling dynamics.

Trained on high-fidelity simulations of pool boiling, \ours jointly models saturated and subcooled boiling across three fluid classes (dielectrics, refrigerants, and cryogens), resolving failure modes of prior models on extreme fluids. 
We show that expert routing exhibits coherent spatial structure and specialization without explicit supervision. 
Quantitatively, \ours matches or exceeds baselines while maintaining physical consistency across heterogeneous boiling configurations. 
We also show zero-shot and few-shot generalization capabilities on downstream tasks such as a new fluid (Opteon 2P50 developed for immersion cooling).
These results demonstrate that mixture-of-experts models are a scalable pathway toward unified surrogate modeling of boiling dynamics and lay the groundwork for broader generalization across scientific ML.

\end{abstract}




 \begin{teaserfigure}
   \centering
   \includegraphics[width=0.9\textwidth]{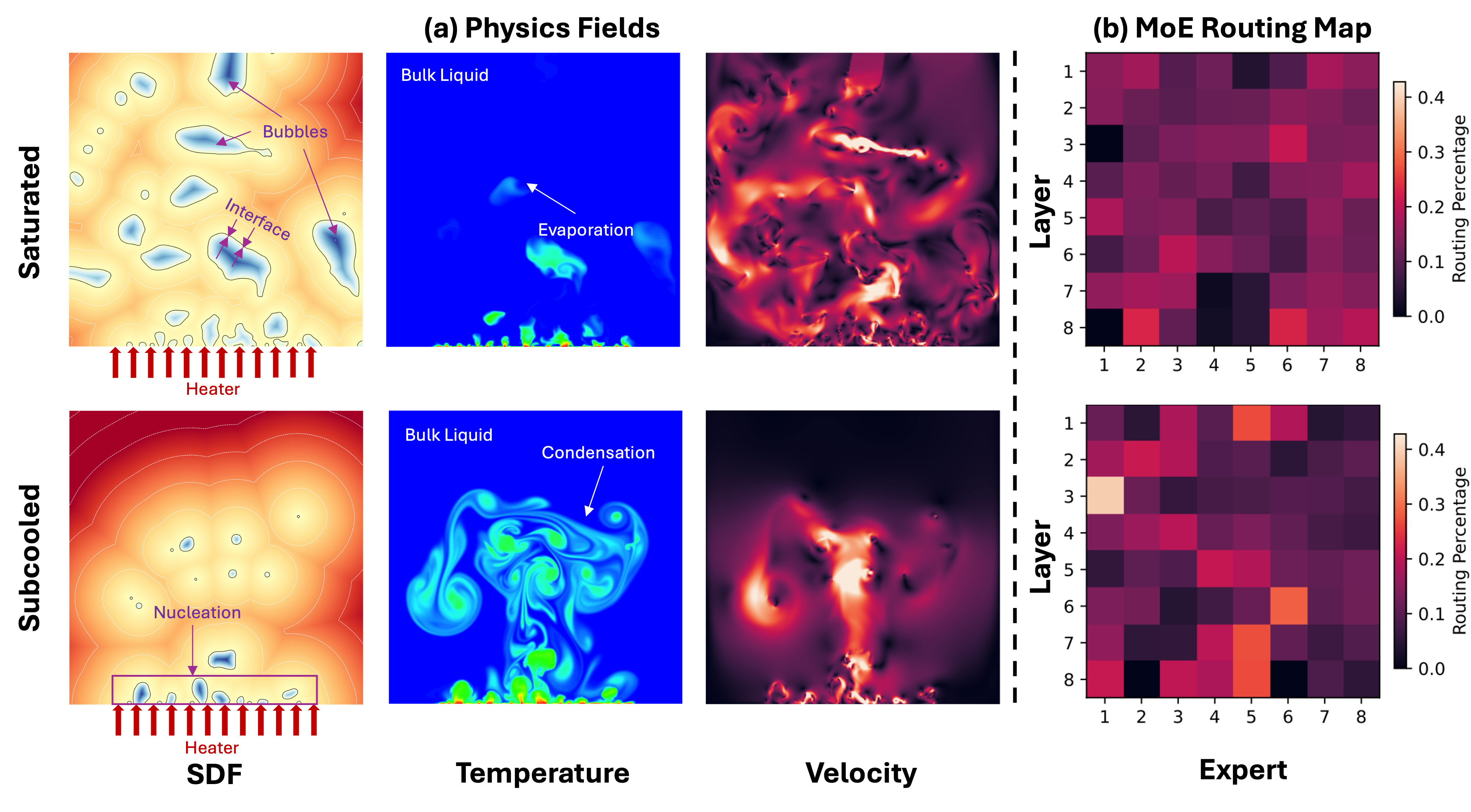}
    \caption{\ours unifies saturated and subcooled boiling across multiple fluids within a single architecture. (a) Physical fields reveal different dynamics: saturated boiling (top) shows concentrated evaporation and rising large bubbles, while subcooled boiling (bottom) exhibits bulk condensation resulting in turbulent vortices and smaller bubbles. (b) Mixture-of-experts (MoE) routing patterns show emergent specialization suggesting learned decomposition of these distinct transport phenomena.}
   \label{fig:physics}
 \end{teaserfigure}

\maketitle


\input{source/intro}
\input{source/background}
\input{source/methods}
\input{source/experiments}
\input{source/related}
\input{source/limitations-and-ethics}
\input{source/conclusion}


\begin{acks}
This work was supported by the Multidisciplinary University Research Initiative (MURI) program by the Office of Naval Research (ONR) under Grant No. N000142412575. We also sincerely thank the Research Cyberinfrastructure Center at the University of California Irvine, for the GPU computing resources on the HPC3 cluster.
\end{acks}

\clearpage
\bibliographystyle{ACM-Reference-Format}
\bibliography{main}

\appendix
\onecolumn

\input{appendix/boiling}
\input{appendix/addn-results}
\end{document}

%% file: source/intro.tex
\section{Introduction}

Boiling is the most efficient known form of heat transfer, in which vapor formation at a heated surface removes thermal energy as bubbles rise and detach. This two-phase cooling phenomenon underpins emerging technologies ranging from data-center thermal management to nuclear energy systems \cite{dirker2019thermal}. 
Growing GPU power densities and rising energy costs have intensified interest in high-performance liquid cooling solutions \cite{azarifar2024liquid}.
However, boiling dynamics arise from tightly coupled multiphase transport, interface motion, and heat transfer, resulting in highly nonlinear behavior.
As bubbles detach and rise from the heater surface, they generate complex vortex structures and turbulent fluctuations in their wakes. 
The resulting velocity field exhibits high-frequency temporal variations and fine-scale spatial structures that are difficult to model. 
The design space is also vast: differences in working fluid, bulk temperature, gravity, heater surface, or geometry can fundamentally alter bubble dynamics, flow patterns, and ultimately cooling efficiency. 
Experimental measurements provide only partial observability of these processes \cite{bucci2016mechanistic}, while high-fidelity simulations remain computationally expensive \cite{dhir2013numerical, sato2018pool, dhruv2019formulation}, motivating data-driven surrogates.

Recent learning-based approaches have shown promise for modeling boiling dynamics from simulation data \cite{hassan2023bubbleml, khodakarami2025mitigating}. 
Yet, existing models are typically trained separately for individual conditions or fluids. 
Attempts to train unified models jointly across heterogeneous boiling problems often exhibit leakage (e.g., hallucinated condensation "wisps" in saturated boiling) and fail under large variations in thermophysical properties, particularly for extreme fluids such as liquid nitrogen (LN2) \cite{hassan2025bubbleformer}. 
As a result, current methods produce collections of specialized models and lack a unified representation of boiling physics.

In this work, we aim to learn a \emph{single model for boiling dynamics} that encodes shared structure while specializing across fundamentally different physical behaviors.
Generalization in scientific ML requires architectural awareness of  when physics fundamentally changes. 
We show that mixture-of-experts (MoE) architectures naturally align with this setting, exhibiting emergent separation of fluid- and condition-dependent dynamics while preserving shared representations.
Combined with physics-informed inductive biases, this results in a single model capable of robust generalization across boiling conditions, where popular conditioning methods (e.g., FiLM \cite{perez2018film}) fail. 
Our main contributions are:

\textbf{1. Unified boiling surrogate.} We present \ours\footnote{\ours code is available at \url{https://github.com/therml-ai/NUCLEUS}}, a single mixture-of-experts model that jointly learns saturated and subcooled pool boiling across multiple fluids with widely varying thermophysical properties.

\textbf{2. Physics-informed architectural design.} We combine neighborhood attention for local transport, signed distance field reinitialization for interface correctness, and MoE routing to enable specialization across phase-change and flow physics.

\textbf{3. Interpretability of expert specialization.} We show expert routing exhibits coherent spatial structure and emergent specialization to physical phenomena without supervision, providing insight into how boiling dynamics are decomposed within the model.

\textbf{4. State-of-the-art accuracy with a unified backbone.} \ours matches or exceeds  baselines while preserving physical consistency over long rollouts in a unified model, demonstrating that MoE-based surrogates can achieve competitive performance. 

\textbf{5. Generalization to downstream tasks.} We demonstrate transfer to a new fluid (Opteon 2P50) unseen during pretraining, highlighting the model's ability to act as a foundation for multiphase multiphysics modeling with minimal finetuning.

%% file: source/background.tex
\section{Background}
\label{sec:background}

\subsection{Problem Formulation}
Boiling is a nonlinear multiphase process governed by the incompressible Navier-Stokes equations coupled with an energy transport equation, solved in both liquid and vapor phases on a spatial domain $\Omega \subset \mathbb{R}^2$ evolving over time $\mathbb{T} = [0, \infty)$.
The liquid and vapor phases $\Omega_L, \Omega_V \subseteq \Omega$ are separated by a dynamically evolving interface $\Gamma$ where phase change occurs. 
We use high-fidelity Flash-X boiling simulations \cite{Dubey_2022, dhruv2019formulation} to solve these equations and generate the BubbleML datasets \cite{hassan2023bubbleml, hassan2025bubbleformer}. We also generate new data modeling the dielectric fluid Opteon 2P50.

At each timestep, the system is represented by three spatially-resolved fields (Figure \ref{fig:physics}): 
(i) temperature $T(x, t) : \Omega \times \mathbb{T} \rightarrow \mathbb{R}$, 
(ii) velocity $U(x, t) : \Omega \times \mathbb{T} \rightarrow \mathbb{R}^2$, and 
(iii) signed distance function $\phi(x, t) : \Omega \times \mathbb{T} \rightarrow \mathbb{R}$, which tracks the phase boundary ($\phi < 0$ in $\Omega_L$, $\phi > 0$ in $\Omega_V$, and $\phi = 0$ at $\Gamma$). 
Together, these define the state $S = (T, U, \phi)$.

In addition to the fields, the dynamics depend on fluid-specific thermophysical properties and operating conditions, collectively denoted $p$. These include non-dimensional numbers (Reynolds Re, Prandtl Pr, Stefan St), thermal conductivity, gravity, and bulk liquid temperature. A full list is in Appendix \ref{app:bubbleml}. These parameters vary significantly across working fluids.

Our learning task is to approximate the forward evolution of the state with a timestep of $\Delta t$: $S(\cdot, t + \Delta t) := M(S(\cdot, t), p)$. 
The model $M$ is applied autoregressively to generate longer trajectories: $S(\cdot, t + \Delta t n) := M^n(S(\cdot, t))$. 
This problem is challenging because boiling combines stochastic nucleation, turbulent transport, moving interfaces, and localized phase change, resulting in chaotic dynamics and strong spatiotemporal heterogeneity.
Long trajectories are inherently sensitive to small perturbations, making pointwise prediction ill-posed. Instead, physically meaningful evaluation relies on statistical field properties and distributions \cite{pope2000turbulentflows}.

\subsection{Physics Heterogeneity and ML Challenges}
\textbf{Subcooled vs. Saturated Boiling.} 
Boiling is categorized by whether the bulk liquid is at (saturated) or below (subcooled) its saturation temperature \cite{faghri2019fundamentals}. 
In saturated boiling, the surrounding liquid is already at its boiling point, so vapor bubbles grow at the heated surface and rise through a nearly isothermal liquid.
In subcooled boiling, the bulk liquid is colder than saturation, causing vapor bubbles to partially or fully condense as they rise.

While both are governed by the same advection–diffusion energy equation,
\begin{equation} 
\label{eq:energy-temp}
\frac{\partial T}{\partial t} + \vec U \boldsymbol{\cdot}  \nabla T = \nabla \boldsymbol{\cdot} \Big[\frac{\alpha'}{\text{Re}\:\text{Pr}}\nabla T \Big] + {S}^{\Gamma}_T
\end{equation}
they exhibit different physical phenomena driven by the \emph{sign} and \emph{spatial distribution} of the phase-change source term ${S}^{\Gamma}_T$, which represents latent heat exchange.
In saturated boiling, liquid evaporates at the heated surface, corresponding to a positive interfacial mass flux (mass leaves liquid) and a negative temperature source ${S}^{\Gamma}_T < 0$. 
In contrast, subcooled boiling is characterized by bulk condensation where vapor collapses within colder liquid, reversing the mass flux (mass enters liquid) and ${S}^{\Gamma}_T > 0$ injecting latent heat back into the surrounding flow.
This leads to distributed thermal sources, condensation-induced turbulent vortices, and fundamentally different transport behavior as seen in Figure \ref{fig:physics}.
This physics-driven heterogeneity motivates our mixture-of-experts design to enable learned separation between physics (e.g., evaporation-dominated vs condensation-dominated) within a unified architecture. 

\textbf{Working Fluids.} 
These differences are compounded by the extreme variations across working fluids in boiling. 
Different fluids (e.g., dielectric coolants, refrigerants, cryogens) have distinct physical properties such as boiling point, surface tension, viscosity, and latent heat which fundamentally alter bubble dynamics and flow patterns.
For example, the boiling point of LN2 (cryogen) is -196\deg C vs 58\deg C for FC-72 (dielectric), a 250\deg C+ range (Appendix Table \ref{tab:fluid_properties}). 
Prior work \cite{hassan2025bubbleformer} successfully unified FC-72 and R515B using a feature-wise linear modulation (FiLM) layer \cite{perez2018film}, but this conditioning fails when the training dataset is extended to include LN2. 
Standard conditioning assumes smooth interpolation across conditions and are unable to handle discrete phase change reversals or extreme property variations.

To date, robust generalization on one of these axes (saturation condition or working fluid) has not yet been achieved, let alone joint modeling across both.  

%% file: source/methods.tex
\section{Methods}

\begin{figure*}[h]
  \vspace{-1em}
  \includegraphics[width=0.9\textwidth]{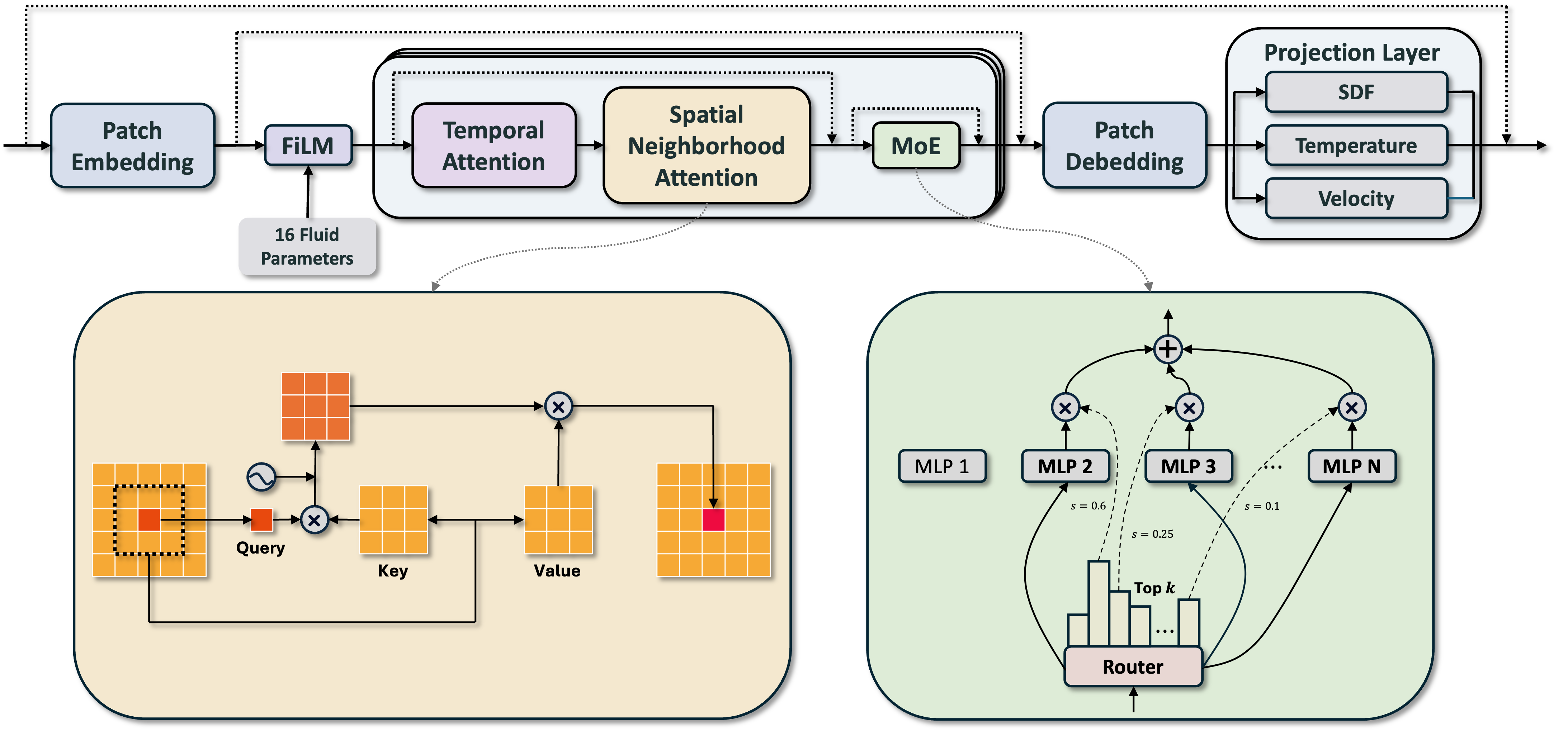}
  \caption{\ours Architecture. Spatiotemporal patches of the state $S=(T, U, \phi)$ are input to a transformer backbone with temporal attention followed by spatial neighborhood attention, enforcing locality aligned with physical interactions. Fluid-specific parameters are input via FiLM conditioning. MoE routes patches to top-k MLP experts, enabling learned specialization of phase-change behaviors. The model predicts the next timestep $S(\cdot, t + \Delta t)$ through field-specific projection heads.}
  \label{fig:nucleus-arch}
\end{figure*}

We introduce \ours, a mixture-of-experts architecture designed to address the heterogeneity challenges outlined in the previous section. 
Our design incorporates three key components motivated by the physics of boiling: (1) \textbf{neighborhood attention} to respect the fine propagation speeds of thermal and momentum transport, (2) \textbf{mixture-of-experts layers} to enable learned specialization across saturation conditions and fluid properties, and (3) \textbf{signed distance field reinitialization} to maintain interface consistency over long rollouts. 
Figure \ref{fig:nucleus-arch} illustrates the overall architecture.

\subsection{Neighborhood Attention}

Physical transport in boiling is governed by local interactions: the state at position $x$ at time $t+\Delta t$ depends only on a bounded neighborhood around $x$ at time $t$, which is determined by fluid velocities and diffusion rates. For example, the top-left corner of a domain has no effect on the bottom-right corner over short-enough time intervals. We found that using relatively small time intervals $\Delta t$ significantly improved results. Bubble nucleation occurs on relatively small timescales, so the model attempting to ``skip'' too many timesteps makes it challenging to learn nucleation. 

Many architectures are not designed to take advantage of this explicit locality and apply global operations. Examples include Fourier Neural Operators (FNO) \cite{li2021fourierneuraloperatorparametric} and vision transformers using dense spatial attention \cite{dosovitskiy2020image}. This forces the model to \emph{learn} that distant regions are irrelevant.
We adopt neighborhood attention \cite{hassani2023neighborhood}, restricting each spatial query to attend only to keys within a fixed radius. This design encodes the inductive bias of the finite domain of influence between timesteps while also dramatically reducing the computational cost necessary for scaling to high-resolution data.
It has also recently been demonstrated to be  effective for weather forecasting \cite{kossaifi2026demystifying}.

\begin{figure}[h]
     \centering
     \begin{subfigure}[b]{0.5\linewidth}
         \centering
         \includegraphics[width=\linewidth]{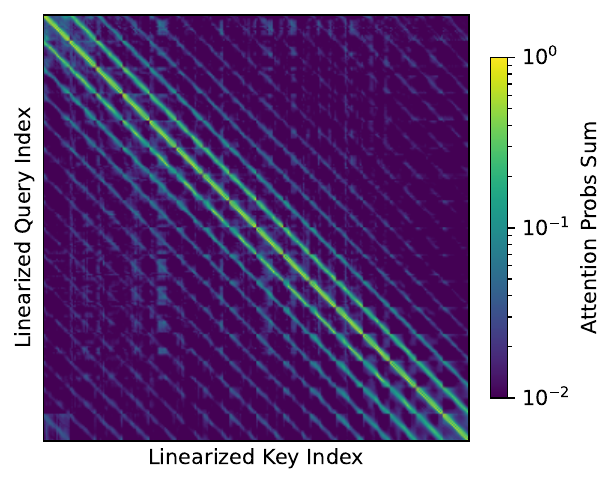}
         \caption{Learned attention pattern $\sum_{h \in [\text{heads}]}\text{Softmax}(Q_hK_h)$ using dense spatial attention (log-scale colormap).}
         \label{fig:attention-heads-sum}
     \end{subfigure}
     \hfill
     \begin{subfigure}[b]{0.43\linewidth}
         \centering
         \includegraphics[width=\linewidth]{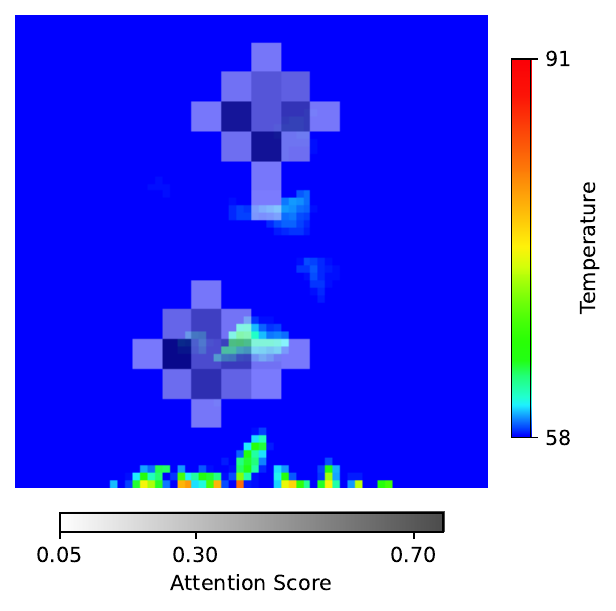}
         \caption{Attention scores for two query patches overlaid on saturated FC-72 frame at 91 \deg C.}
         \label{fig:attention-query-overlay}
     \end{subfigure}
     \caption{Empirical validation of ViT-style global attention prioritizing local regions. (a) Across all heads, attention learns to prioritize local regions; revealing that long-range interactions provide little benefit.  (b) Example queries shows attention scores $>0.05$ are localized around the query position, consistent with physics-based transport.}
     \label{fig:attention-analysis}
\end{figure}

Figure~\ref{fig:attention-heads-sum} shows the sum of attention probabilities for each head of the first attention block in a model trained using standard dense spatial attention. The model learns a noisy approximation of neighborhood attention, revealing that global receptive fields provide little benefit.
Figure~\ref{fig:attention-query-overlay} shows example attention scores for selected query locations, again demonstrating the localized receptive fields and validating our design.

\subsection{Mixture of Experts}

\begin{figure*}[h]
    \centering
    \includegraphics[width=0.9\linewidth]{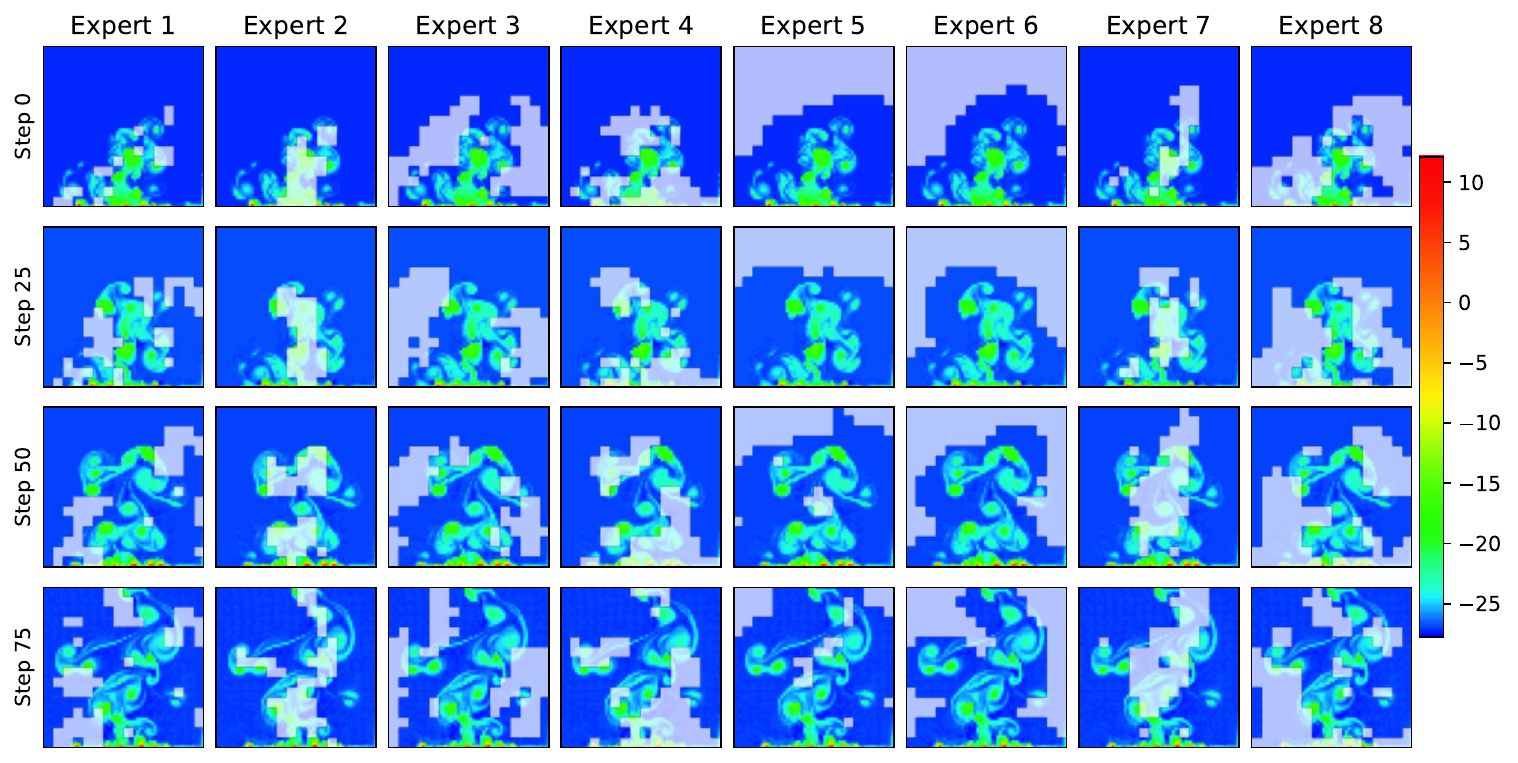}
    \caption{Spatiotemporal expert specialization in subcooled R515B boiling. Temperature fields at four timesteps overlaid with routing patterns from the first MoE layer of \ours (8 experts, top-2 routing). Grey patches indicate regions routed to each expert. Experts 2 and 7 tends to receive patches from bubbles or hotter regions, while Expert 6 specializes in bulk liquid.}
    \label{fig:moe-routing-overlay}
\end{figure*}

The BubbleML dataset is a union of several categories of real-world boiling. 
For instance, it includes saturated boiling of liquid nitrogen at heater temperature ranges from -191 to -159\deg C and subcooled boiling of FC-72 with heater temperatures ranging from 75 to 107\deg C. 
Note that the heater temperatures are intended to stay in the nucleate boiling regime.
As discussed in Section \ref{sec:background}, different combinations of liquid temperature, heater temperature, and working fluid result in fundamentally different dynamics. 
In the past, this extreme heterogeneity has made it difficult to train a single unified model on the entire dataset that works well for each of the individual types of boiling \cite{hassan2025bubbleformer}. 

We address this challenge by replacing all of the dense MLP layers with sparse MoE layers \cite{shazeer2017}.
Each MoE layer contains $E=8$ experts (feed-forward MLPs) and a learned router that computes routing probabilities for each patch. We use top-$k=2$ routing. Expert outputs are weighted by their routing probabilities and summed. To encourage load balancing across experts, we add an auxiliary load balancing loss \cite{fedus2022switchtransformersscalingtrillion}.

MoE is motivated by the mathematical structure of multiphase transport. Equation \ref{eq:energy-temp} shows that saturation state fundamentally changes the \emph{operator} governing energy transport through the sign and spatial distribution of $S_T^{\Gamma}$. Standard conditioning approaches like FiLM modulate \emph{representations} but preserve a single computational pathway. MoE instead allows the model to route to different pathways, effectively learning separate approximate operators for evaporation-dominated vs. condensation-dominated transport. This architectural choice aligns with the physics: we expect fundamentally different solution structures.

Figure~\ref{fig:moe-routing-overlay} reveals emergent spatial and temporal structures in expert routing, validating our design hypothesis. 
Across 75 timesteps spanning bubble nucleation, growth, and detachment, experts maintain consistent physical specializations.
Experts 2 and 7 concentrate on bubbles and high-temperature regions above the heater, tracking bubbles as they rise. Expert 6 preferentially receives patches in the bulk liquid region.
Expert 1 specializes in thermal boundary layers and regions with strong temperature gradients.
This emergent learned specialization to distinct physical phenomena suggests that the model learns to decompose the heterogeneous problem along physically meaningful axes, notably \emph{without} explicit supervision.

\subsection{SDF Reinitialization}

\begin{figure}[h]
    \centering
    \includegraphics[width=0.8\linewidth]{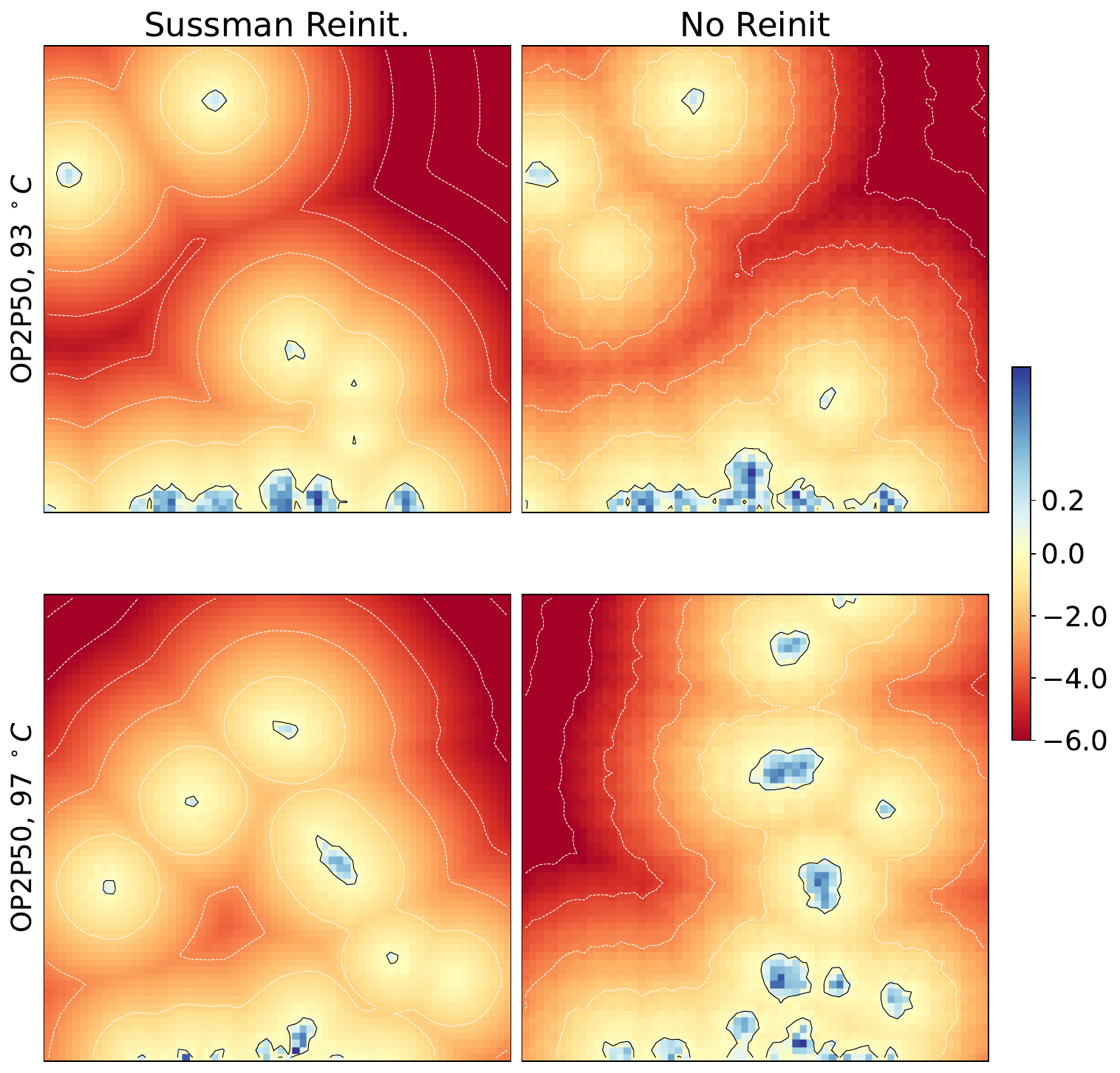}   
    \caption{Compares the signed distance function $\phi$ contours during rollouts without (top) and with (bottom) reinitialization. Spurious "wave" artifacts appear without reinitialization as errors accumulate far from bubble interfaces. Reinitialization restores the Eikonal property. 
    }
    \label{fig:sdf-reinit-compare}
    \vspace{-1em}

\end{figure}

One issue we observed empirically with the local models is that the signed distance function $\phi$ develops errors in regions far from bubble interfaces during autoregressive inference. For example, when bubbles are sparse or concentrated near the heater, the upper portion of the domain may lie near the edge of the model's receptive field, preventing interface dynamics from being fully propagated. As a result, $\phi$ can drift from the Eikonal equation $||\nabla\phi||_2 = 1$ \cite{osher2003level}. Nevertheless, we found that the SDF remains sufficiently accurate near the bubble interfaces.

To address this deterioration, we apply the Sussman method to reinitialize the SDF in regions far from the interfaces during inference \cite{osher2003level, sussman1999reinit}. The Sussman method is an iterative scheme that solves $\frac{\partial \phi}{\partial \tau} + \text{sgn}(\phi_0)\left(|\nabla \phi| - 1\right) = 0$ using boundary conditions defined by the zero level set (the predicted interface where $\phi=0$), which remains accurate within the model's receptive field.

During inference, we apply the Sussman reinitialization after each step.  In practice, only a few iterations are required to substantially improve SDF stability. Figure \ref{fig:sdf-reinit-compare} compares the SDF $\phi$ with and without reinitialization, demonstrating that this correction alleviates error accumulation during autoregressive rollouts. Although we use this as a test-time correction to improve rollout quality, it can be incorporated into training since each step is differentiable.

\subsection{Training}

We use a relative $L_1$ loss to balance the wide range of magnitudes across physical fields: $\mathcal{L}_f = \frac{\lVert f_{\mathrm{model}} - f_{\mathrm{target}} \rVert_1} {N_f / N_{\max}}$, $f \in \{\phi, T, U_x, U_y\}$,
$N_{\max} = \max(N_f) \nonumber$.
For the three fields $\{\phi, U_x, U_y\}$, we compute $N_f = \lVert f_{\mathrm{target}} \rVert_1$.
The only exception is the temperature, for which we subtract the bulk liquid temperature in the norm: $N_T =\lVert T_{\mathrm{target}} - T_{\mathrm{bulk}} \rVert_1$. 
Without this, liquid nitrogen, which can have a bulk temperature of $-204$\deg C, would artificially have a much smaller loss than R515B, which is more closely centered at 0\deg C. 
Each field is normalized by the scaling factor $N_f / N_{\max}$ to balance their relative contributions.
The total loss is the sum of the four field losses $\mathcal{L} = \sum_f \mathcal{L}_f$.

We normalize by $N_{\max}$ because of the MoE load balancing loss. Since the purpose of the relative loss is to equally weight the losses for the different fields, we wanted to keep the weighted loss in a similar range to an unweighted loss during ablation studies. This simplified experimentation with different MoE load balancing losses, allowing us to compare weighted and unweighted losses without performing additional sweeps of load balancing loss weights.

To encourage uniform expert utilization, we apply the Switch Transformer auxiliary
load balancing loss from \cite{fedus2022switchtransformersscalingtrillion}. For a batch of $T$ tokens routed to $E$ experts with
top-$k$ selection, the loss is
$\mathcal{L}_{\mathrm{LB}} = \alpha E \sum_{e=1}^{E} f_e \, \bar{p}_e$,
where
$f_e = \frac{1}{Tk} \sum_{x} \sum_{j=1}^{k} \mathbf{1}\{g_j(x)=e\}$,
$\bar{p}_e = \frac{1}{T} \sum_{x} p_e(x)$,
and $p(x)=\mathrm{softmax}(\mathrm{router}(x))$.
$\alpha$ is a hyperparameter for the auxiliary loss weight and is set to 0.01 across all experiments.

All models and ablations are on an NVIDIA A30 GPU with 24GB HBM and batch size of $4$.
Each model is trained for up to $40$ epochs, and we select the model that performed best on a hold-out evaluation set.
We use a cosine annealing learning rate scheduler with $5000$ warmup iterations and a peak learning rate of $10^{-3}$ that decays to $10^{-6}$. During training, we apply two simple data augmentations: random horizontal flips and additive zero-mean Gaussian noise with randomly sampled magnitude to improve robustness during autoregressive inference.

%% file: source/experiments.tex
\section{Experiments and Results}

We evaluate \ours on multiple axes: (1) \textbf{accuracy} across all fluid-condition combinations, (2) \textbf{stability} over autoregressive rollouts needed for statistical evaluation of turbulent dynamics, (3) \textbf{generalization} to unseen new fluid, and (4) \textbf{performance} efficiency of training and inference across different resolutions.

\subsection{Setup}

\paragraph{\textbf{Dataset.}}
We use the BubbleML pool boiling dataset consisting of 100+ boiling configurations varying in subcooling, heater temperature, and fluids to train \ours. The dataset covers three real-world fluids (FC72, R515B, LN2) and two types of boiling (saturated and subcooled) with vastly different properties.

\paragraph{\textbf{Physical Metrics.}} Classical boiling research has primarily focused on mean quantities such as the heat transfer coefficient (HTC) and critical heat flux (CHF), which collapse complex multiphase dynamics into low-dimensional performance metrics \cite{rohsenow1952method, zuber1959hydrodynamic}.
This arose largely from historical measurement and modeling limitations, where experiments provided access mainly to wall temperature and net heat flux, while accurately resolving evolving interfaces and flow fields remained intractable \cite{bucci2016mechanistic, liu2020uncertainty}.  
In modern applications such as GPU and data center cooling, failures are often driven by localized overheating and transient dryout rather than degradation in mean HTC alone, motivating analysis beyond aggregate metrics.

Because \ours predicts full temperature, velocity, and phase fields, we evaluate performance using statistical properties of these fields rather than pointwise comparisons to a single numerical realization. 
This is consistent with standard practice in boiling studies, where individual bubble trajectories are stochastic and reproducibility is assessed through averaged quantities and distributions \cite{faghri2019fundamentals}. 
Temperature statistics are particularly important because thermal transport directly determines wall superheat and HTC. Accordingly, we report the spatiotemporal mean temperature $\langle T \rangle_{x,y,t}$.

To characterize the transport mechanisms underlying heat removal, we additionally measure time-averaged velocities in the liquid, vapor, and interface regions. We also analyze temperature and velocity distributions, which capture nucleation activity and vapor removal rates. Together, these statistics provide mechanistic insight into how cooling performance emerges from multiphase dynamics, complementing aggregate metrics while enabling field-level evaluation of model behavior.

\paragraph{\textbf{Baselines.}}
We compare \ours against several transformer-based PDE surrogates, including architectures specifically designed for boiling as well as more general foundation models for PDEs. Prior work has shown that UNet-style architectures and FNO variants perform poorly on this problem \cite{hassan2023bubbleml}, so we exclude them from evaluation. 
For boiling-specific modeling, we compare against \emph{Bubbleformer} \cite{hassan2025bubbleformer}, a transformer with axial attention. Axial attention has similar issues to global attention: processing the full extent of each axis is inefficient for boiling systems where dynamics are dominated by local interactions over short time ranges. 
To evaluate against broader PDE surrogate architectures, we additionally compare with \emph{MoE-DPOT} \cite{wang2025mixtureofexperts} and \emph{Poseidon} \cite{herde2024poseidon}. 
MoE-DPOT is a MoE architecture that routes entire inputs, rather than local patches, to experts and uses a class token to distinguish between different input categories. Poseidon is a hierarchical multiscale transformer designed to be a foundation model for PDEs \cite{herde2024poseidon}. 

We additionally ablate several design choices in \ours. Specifically, we evaluate: (a) spatial attention, replacing neighborhood attention with axial attention and global (ViT-style) attention, (b) MoE versus standard dense MLP layers, and (c) Sussman test-time reinitialization versus an Eikonal loss (Appendix \ref{app:ablation-sdf}).

\subsection{Single-step Accuracy}
\begin{figure*}[h]
    \centering
    \includegraphics[width=\linewidth]{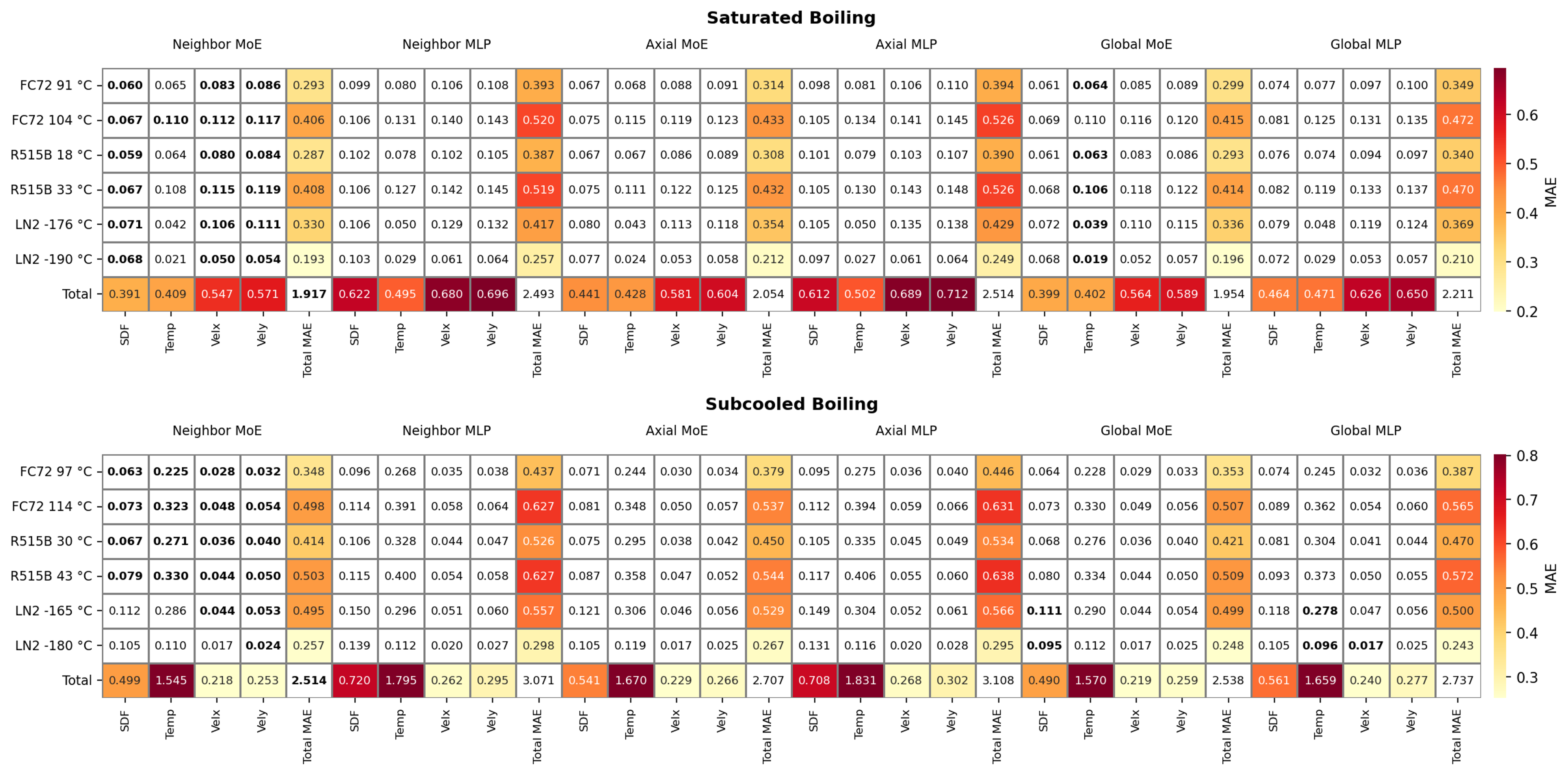}
    \caption{Mean absolute error (MAE) heatmaps for \textbf{(top)} saturated boiling and \textbf{(bottom)} subcooled boiling. Rows correspond to liquid conditions and columns report MAE for the signed distance field, temperature, x-velocity, y-velocity, and aggregate MAE across all fields. The bottom row aggregates errors across all conditions. Across both boiling types, Neighbor MoE achieves the lowest total MAE and outperforms MLP baselines, demonstrating robust accuracy across heterogeneous boiling physics.}
    \label{fig:heatmap_boiling}
    \vspace{-1em}
\end{figure*}

We evaluate the single-step prediction accuracy of all models.
Because the time-interval is sufficiently short, turbulent divergence is minimal, allowing for a direct field-wise comparison between model predictions and high-fidelity numerical simulations. 
Figure \ref{fig:heatmap_boiling} shows that across all boiling types and operating conditions, \emph{MoE models consistently outperform their dense MLP baselines}. 
Although Global and Axial MoE improve upon their respective MLP baselines, Neighbor MoE matches or outperforms their performance, suggesting that fully global receptive fields are not required for attending to localized physical cues. 
Errors are consistently higher in subcooled boiling, reflecting the increased complexity introduced by strong thermal gradients, condensation, and re-entrant interface dynamics.
These conditions further amplify the performance gap between MoE and dense architectures. 

Table \ref{tab:bubbleformer-comp} summarizes the temperature MAE for the baselines. Model parameter counts are reported in Appendix \ref{app:parameters}.
Although MoE-DPOT and Poseidon achieve competitive single-step prediction errors, qualitative evaluation reveals notable deficiencies in spatial fidelity. As shown in Appendix \ref{app:viz-baselines}, both models produce visually degraded interfaces and thermal structures even at single-step inference. During rollout, these errors accumulate rapidly, leading to severe deterioration in dynamics. 
In contrast, \ours preserves sharper thermal structures and more consistent multiphase dynamics. Bubbleformer performs significantly worse overall, exhibiting catastrophic failures when trained on the unified dataset. This is consistent with prior observations \cite{hassan2025bubbleformer} where features from incompatible boiling are spuriously transferred (e.g., subcooled condensation appearing in saturated boiling).

\begin{table}[h]
    \centering
    \caption{Mean absolute error (\deg C) of the temperature field vs. ground-truth simulation. Bubbleformer (BFormer) fails on the unified dataset with large errors on extreme fluids. MoE-DPOT and Poseidon (ScOT) are more competitive.}
    \begin{tabular}{l|rrrr}
        \toprule
        Case  & BFormer & ScOT & MoE-DPOT & \textbf{NUCLEUS}\\
        \midrule
        Sat FC72 91\deg C & 74.723 & 0.424 & 0.159 & \textbf{0.143} \\
        Sat FC72 104\deg C & 74.854 & 0.527 & 0.269 & \textbf{0.246} \\
        Sat R515B 18\deg C & 0.183 & 0.351 & 0.140 & \textbf{0.126} \\
        Sat R515B 33\deg C & 0.274 & 0.434 & 0.239 & \textbf{0.223} \\
        Sat LN2 -176\deg C & 165.428 & 0.930 & 0.117 & \textbf{0.076} \\
        Sat LN2 -190\deg C & 165.487 & 0.542 & 0.033 & \textbf{0.028} \\
        \midrule
        Sub FC72 97\deg C & 70.250 & 0.938 & 0.631 & \textbf{0.448} \\
        Sub FC72 114\deg C & 70.595 & 0.954 & 0.792 & \textbf{0.733} \\
        Sub R515B 30\deg C & 1.516 & 0.726 & 0.671 & \textbf{0.564} \\
        Sub R515B 43\deg C & 1.562 & 0.829 & 0.774 & \textbf{0.725} \\
        Sub LN2 -165\deg C & 154.764 & 1.060 & 0.671 & \textbf{0.532} \\
        Sub LN2 -180\deg C & 155.715 & 1.116 & 0.377 & \textbf{0.177} \\
        \bottomrule
    \end{tabular}
    \label{tab:bubbleformer-comp}
    \vspace{-1em}
\end{table}

\subsection{Trajectory Rollouts}

\begin{figure*}[h]
    \centering
    \includegraphics[width=0.98\linewidth]{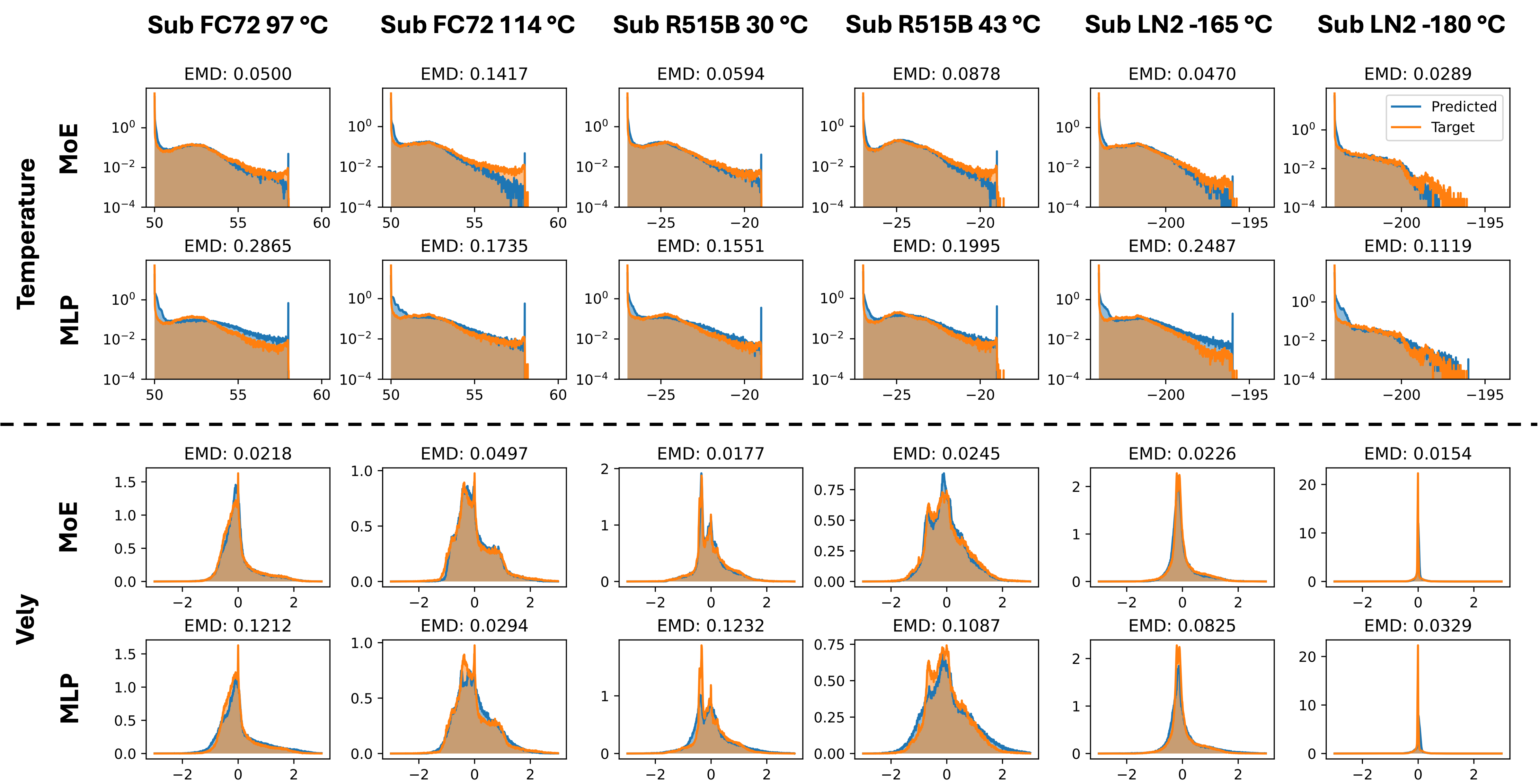}
    \caption{Distribution of temperature and vertical velocity for subcooled boiling over 100 timesteps autoregressive rollout. Shown are the empirical distributions from ground-truth simulations and model predictions. Agreement is quantified using Earth Mover’s Distance (EMD), which captures deviations in both mean behavior and asymmetric tail structure.}
    \label{fig:subcooled-liquid-temp-dist}
\end{figure*}

We evaluate long-horizon autoregressive behavior by comparing 
the \emph{distributional statistics} of key physical fields between predicted and ground-truth trajectories. 
Figure \ref{fig:subcooled-liquid-temp-dist} shows the empirical distributions of liquid temperature and vertical velocity ($U_y$) for subcooled boiling. 
These quantities are directly tied to boiling performance: temperature distributions characterize thermal transport and cooling efficiency, while vertical velocity governs buoyancy-driven bubble rise, detachment, and vapor removal dynamics.

For each rollout, we compute the Earth Mover's Distance (EMD) between predicted and ground-truth distributions. 
The MoE models closely match both temperature and velocity distributions throughout the rollout horizon. Modeling $U_y$ is particularly challenging due to its sensitivity to interface motion and buoyancy-driven acceleration. Nevertheless, MoE models accurately capture both the central mass and the asymmetric tails associated with fast-rising bubbles. 
In contrast, MLP baselines exhibit systematic drift, underestimating tails associated with bubble-induced mixing. 

\subsection{Downstream Task}

We evaluate transferability with a downstream adaptation task on a new working fluid: Opteon 2P50 (OP2P50), a dielectric designed for two-phase immersion cooling with zero ozone depletion potential. The new simulations generated are summarized in Table \ref{tab:op250_conditions}.

\begin{figure*}[h]
     \centering
     \begin{subfigure}[t]{0.19\linewidth}
         \centering
         \includegraphics[width=\linewidth]{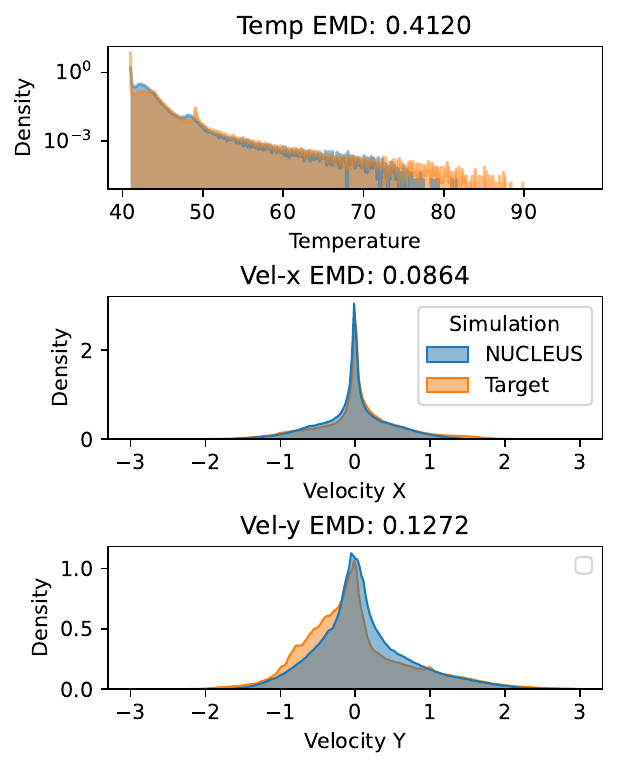}
         \label{fig:op250-rollout-dist}
     \end{subfigure}
     \hfill
     \begin{subfigure}[t]{0.78\linewidth}
         \centering
         \includegraphics[width=\linewidth]{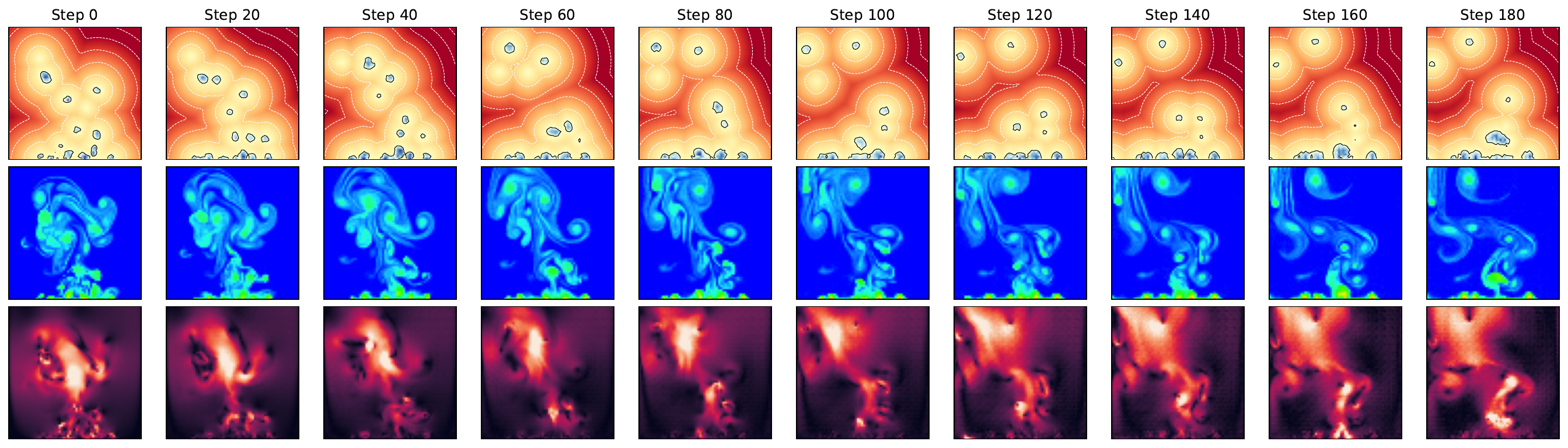}
         \label{fig:op250-rollout-timesteps}
     \end{subfigure}
     \vspace{-1em}
     \caption{Few-shot adaption of \ours to subcooled OP2P50 boiling using only three simulations. Left: Temperature and velocity distributions for an unseen rollout with a heater temperature of $97$\deg C. Right: Example autoregressive rollout after finetuning, demonstrating stable interface, thermal transport, and velocity evolution despite limited finetuning data.}
     \label{fig:op250-rollout}
\end{figure*}

\begin{figure}
    \centering
    \includegraphics[width=1.0\linewidth]{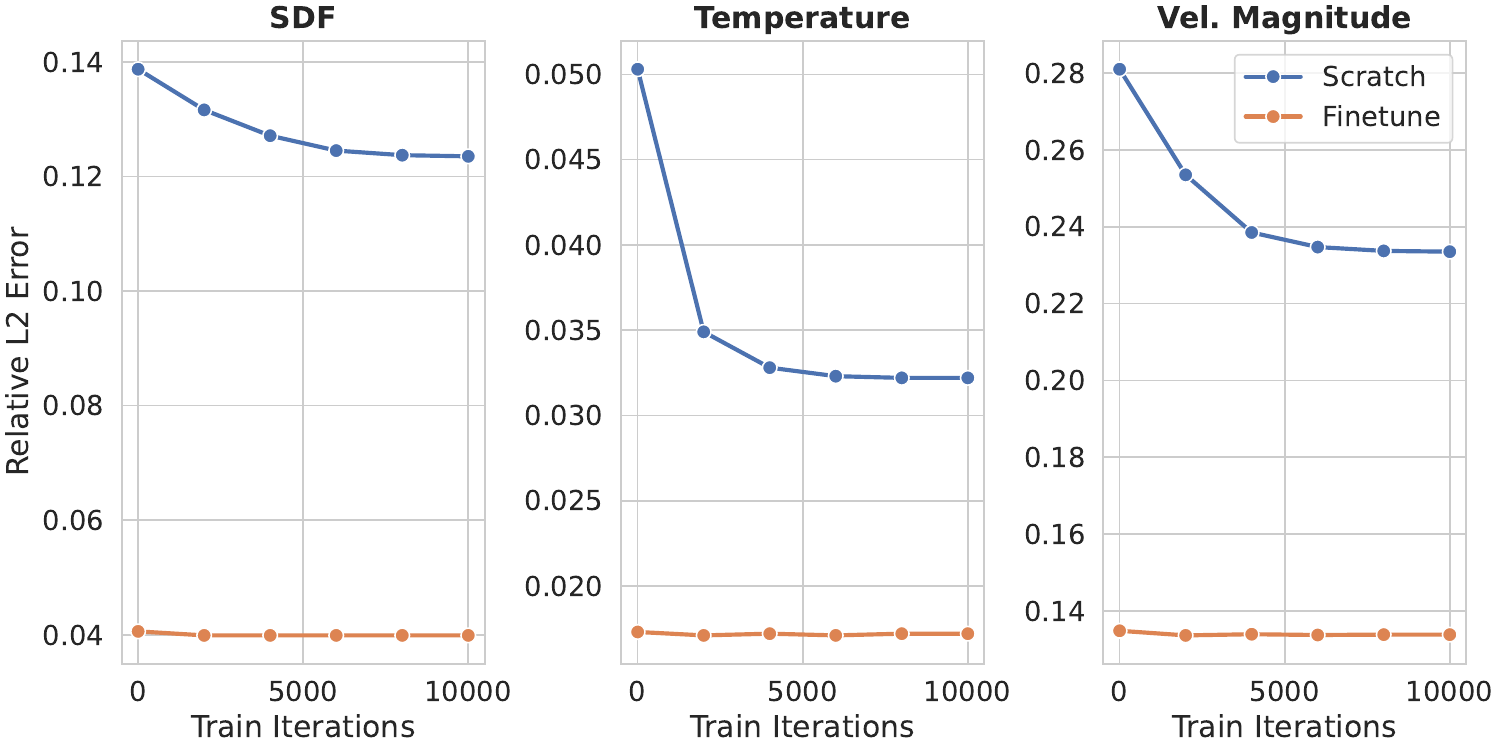}
    \caption{Comparison of training \ours from scratch on OP2P50 versus finetuning a pretrained model. Shown is the single-step relative L2 error for 25 random samples from the OP2P50 test simulations.}
    \label{fig:scratch-vs-finetune-op2p50}
\end{figure}

To assess low-data adaptation, we perform few-shot fine-tuning using only three OP2P50 simulations with heater temperatures of 85, 89, and 101\deg C, reserving 81\deg C for validation. The pretrained \ours model is finetuned for five epochs with a learning rate of $10^{-5}$.  Finetuning only takes minutes. Figure \ref{fig:op250-rollout} shows stable rollout results for heater temperature 97\deg C, demonstrating the benefit of training on a large unified dataset. We further compare few-shot fine-tuning against training directly from scratch on the OP2P50 dataset. As shown in Figure \ref{fig:scratch-vs-finetune-op2p50}, fine-tuning converges substantially faster and achieves lower single-step errors across all physical fields.

\subsection{Performance and Scaling Efficiency}

Both MoE and neighborhood attention are beneficial for scalable boiling surrogates. MoE layers enable larger model capacity, while neighborhood attention enables efficient training and inference at higher resolutions. As shown in Table \ref{tab:compute_efficiency_metrics}, \ours is $\sim 4\times$ faster and requires $\sim 3\times$ less memory than comparable dense MLP baselines. This advantage becomes more pronounced at higher resolution. Table \ref{tab:efficiency_multi_res} reports the costs for a single transformer block across multiple resolutions at a fixed patch size. At $2048\times2048$, \ours is $\sim 5-6\times$ faster and $2.5\times$ more memory efficient than a transformer block using full spatial attention and dense MLP layers.

\input{tables/model_compte_efficiency}
\input{tables/block_scaling}


%% file: tables/model_compte_efficiency.tex
\begin{table}[h]
\centering
\caption{Compute efficiency metrics. All experiments use $64\times64$ resolution inputs, batch size 8, embedding dimension 384, patch size 4, and 12 transformer blocks. MoE models use 8 experts with top-2 routing. The MLP intermediate dimension is chosen to match the total parameters count across models.}
\label{tab:compute_efficiency_metrics}
\setlength{\tabcolsep}{6pt}
\begin{tabular}{lrrrr}
\toprule
\textbf{Model} &
\begin{tabular}[c]{@{}r@{}}\textbf{Active/}\\\textbf{Total (M)}\end{tabular} &
\begin{tabular}[c]{@{}r@{}}\textbf{Infer}\\\textbf{(ms)}\end{tabular} &
\begin{tabular}[c]{@{}r@{}}\textbf{Train}\\\textbf{(ms)}\end{tabular} &
\begin{tabular}[c]{@{}r@{}}\textbf{VRAM}\\\textbf{(MB)}\end{tabular} \\
\midrule
NATTEN + MLP    & 128.1/128.1 & 328.8          & 1010.4          & 15275 \\
Full Attn + MLP & 128.1/128.1 & 329.8          & 1011.6          & 15369 \\
Bubbleformer    & 128.0/128.0 & 369.8          & 1120.3          & 17714 \\
Full Attn + MoE & 43.0/128.0  & \textbf{72.7} & \textbf{208.5} & 5622  \\
NUCLEUS         & 43.0/128.0  & 74.4           & 217.2           & \textbf{5537} \\
NUCLEUS         & 128.0/467.7 & 99.5           & 306.2           & 10503 \\
\bottomrule
\end{tabular}
\vspace{-1em}
\end{table}

%% file: tables/block_scaling.tex
\begin{table}[h]
\centering
\caption{Compute efficiency and training memory usage for a single transformer block at increasing spatial resolutions. All models use patch size 8 and embedding dimension 256. Each MoE expert uses intermediate dimension 1024.}
\label{tab:efficiency_multi_res}
\setlength{\tabcolsep}{7pt}
\begin{tabular}{llrrr}
\toprule
\textbf{Res} &
\textbf{Model} &
\begin{tabular}[c]{@{}r@{}}\textbf{Infer}\\\textbf{(ms)}\end{tabular} &
\begin{tabular}[c]{@{}r@{}}\textbf{Train}\\\textbf{(ms)}\end{tabular} &
\begin{tabular}[c]{@{}r@{}}\textbf{VRAM}\\\textbf{(MB)}\end{tabular} \\
\midrule
128
& NATTEN + MLP            & 2.4   & 5.9    & 116 \\
& Full Attn + MLP         & \textbf{2.1} & \textbf{5.4} & 117 \\
& Full Attn + MoE         & 2.8   & 6.9    & 61 \\
& Bubbleformer            & 2.9   & 6.7    & 136 \\
& \textbf{NUCLEUS (ours)} & 3.1   & 7.3    & \textbf{60} \\
\midrule
512
& NATTEN + MLP            & 15.3  & 46.3   & 1293 \\
& Full Attn + MLP         & 16.1  & 48.4   & 1303 \\
& Full Attn + MoE         & 7.2   & 19.1   & 537 \\
& Bubbleformer            & 19.5  & 56.2   & 1722 \\
& \textbf{NUCLEUS (ours)} & \textbf{6.5} & \textbf{17.1} & \textbf{527} \\

\midrule
2048
& NATTEN + MLP            & 242.7 & 745.5  & 19552 \\
& Full Attn + MLP         & 498.6 & 1659.6 & 19712 \\
& Full Attn + MoE         & 351.8 & 1158.1 & 8178 \\
& Bubbleformer            & OOM   & OOM    & OOM \\
& \textbf{NUCLEUS (ours)} & \textbf{96.3} & \textbf{250.5} & \textbf{8018} \\

\bottomrule
\end{tabular}
\end{table}

%% file: source/related.tex
\section{Related Work}

\textbf{Boiling Heat Transfer: Experiments and Simulations.} 
The study of boiling physics has evolved from empirical correlations to mechanistic modeling to high-fidelity numerical simulations. Early work established the boiling curve relating heat flux to wall superheat \cite{nukiyama1966maximum} and  identified distinct regimes such as nucleate boiling, where discrete bubbles form and depart from heated surfaces \cite{han1962mechanism}. 
Foundational correlations for HTC and CHF were subsequently developed and remain widely used in engineering practice today \cite{rohsenow1952method, zuber1959hydrodynamic}.
As computational capabilities advanced, research shifted toward mechanistic modeling of liquid-vapor dynamics. 
Numerical methods, including volume-of-fluid and level-set approaches, enabled direct simulation of coupled momentum and energy transport with evolving liquid-vapor interfaces \cite{dhir2013numerical, sato2018pool, dhruv2019formulation}. 
These studies revealed the importance of microlayer evaporation, contact line dynamics, and bubble interaction that are not captured by empirical correlations alone. 
However, such simulations remain computationally expensive, often requiring days to weeks for a single operating condition on modern HPC systems.
More recently, large-scale simulation campaigns \cite{Dubey_2022} spanning multiple fluids and thermodynamic conditions have enabled the creation of datasets like BubbleML \cite{hassan2023bubbleml, hassan2025bubbleformer} and MPFBench \cite{shadkhah2025mpfbench}. These datasets have made data-driven surrogate modeling for boiling viable for the first time.

\noindent \textbf{Neural Surrogates for Fluid Dynamics and Multiphase Flow.}
Neural surrogate modeling has emerged as a promising approach for accelerating the solution of PDEs \cite{brunton2020machine, azizzadenesheli2024neural, eyring2024pushing}.
Neural operators \cite{li2021fourierneuraloperatorparametric, poels2024accelerating} and transformer-based surrogates \cite{herde2024poseidon, wu2024transolver, wang2025cross} have shown promise on a variety of advection–diffusion and fluid dynamics problems. 
For boiling, the BubbleML benchmarks showed that FNO variants \cite{tran2023factorized} perform well at predicting velocity but struggle with temperature, especially high-frequency modes near interfaces and condensation trails \cite{hassan2023bubbleml}. 
Subsequent work addressed spectral bias and oversmoothing in neural operators using high-frequency scaling \cite{wang2022antioversmoothing} improving the prediction of condensation vortices in subcooled boiling \cite{khodakarami2025mitigating}.
Bubbleformer demonstrated that transformers with axial attention and FiLM conditioning can predict nucleation and therefore forecast boiling evolution from past states alone \cite{hassan2025bubbleformer}.
UNet-style CNNs and FNOs struggle with accurate SDF prediction due to sharp interface discontinuities and difficulty modeling nucleation events.
However, FiLM conditioning was only shown to unify FC-72 and R515B datasets, while unified modeling across saturated and subcooled boiling or extreme fluids remains challenging.

\noindent \textbf{Mixture-of-Experts for Scientific ML.}
Sparse MoE layers have emerged as a popular alternative to the dense feed-forward layers in transformers. 
They route inputs to a subset of expert networks, scaling model capacity without proportionally increasing compute \cite{shazeer2017,krajewski2024scaling}.
MoE finds applications in LLMs \cite{dai2024deepseekmoe}, vision \cite{riquelme2021scaling}, reinforcement learning \cite{obando2024mixtures}, and machine learning for science \cite{chalapathi2024scaling}.

MoEs are attractive for multi-domain and multi-task settings, where different experts can specialize in distinct data regimes~\cite{fedus2022switchtransformersscalingtrillion}. In scientific ML, MoE-POT \cite{wang2025mixtureofexperts} demonstrated that equation-specific expert routing can improve large-scale PDE surrogate pretraining while resulting in interpretable specialization across equation families. Our work extends these ideas to multiphase flows, where the challenge is to learn distinct representations for physical processes occurring simultaneously across spatial regions and scales. 

%% file: source/limitations-and-ethics.tex
\section{Limitations and Ethical Considerations}

This work focuses on pool boiling and does not model \emph{flow boiling}, where externally driven flow introduces inlet-outlet boundary conditions and flow-regime transitions. Extending these methods to broader classes of multiphase flows remains future work.

Like other autoregressive PDE surrogates \cite{lippe2023pderefinerachievingaccuratelong}, \ours accumulates errors during long rollouts. The primary intended use is to accelerate thermal-fluid simulation and design exploration for energy-efficient cooling systems \cite{congress}. Deploying these models directly in safety-critical applications such as nuclear thermal management requires rigorous validation against high-fidelity simulations or experimental measurements.

%% file: source/conclusion.tex
\section{Conclusion}

\ours learns to model diverse types of pool boiling across multiple fluids and operating conditions while accurately reproducing key distributional properties of two-phase flows.
Ablations validate the architectural choices (neighborhood attention, MoE layers, and SDF reinitialization), and downstream experiments reveal promising few-shot transfer to previously unseen fluids using only limited finetuning data.
To our knowledge, \ours is the first architecture to successfully unify saturated and subcooled boiling across multiple extreme fluids within a single surrogate model.

%% file: appendix/boiling.tex
\section{BubbleML Dataset}
\label{app:bubbleml}
To test the generalization of \ours to physical scenarios not seen during pretraining, we generate out of distribution datasets. Specifically, we perform Flash-X \cite{Dubey_2022} simulations for an industrially popular dielectric fluid, \textbf{Opteon 2P50}, which finds applications in 2-phase immersion cooling of electronic components. We generate 6 different boiling simulations at different wall superheats along the nucleate region of the boiling curve, as shown in Table \ref{tab:op250_conditions}. The subcooling is set to 8°C. The nucleation site density is assumed to increase linearly with increasing heater temperature. We release the newly generated Op2p50 datasets on Hugging Face\footnote{Newly generated data is available at \url{https://huggingface.co/datasets/hpcforge/BubbleML_2}} as an extension to BubbleML 2.0 \cite{hassan2025bubbleformer}.

\begin{table}[htbp]
\centering
\caption{Boiling curve data for simulations of dielectric fluid Opteon 2P50 used in finetuning. Saturation temperature is in non-dimensional units $(T_\text{sat} - T_\text{bulk}) / (T_\text{wall} - T_\text{bulk})$.}
\label{tab:op250_conditions}

\small
\begin{tabular}{c|c|c|c|c|c}
\hline
\textbf{Working Fluid} 
& \textbf{Wall Temp (°C)} 
& \textbf{Nucleation Sites} 
& $\boldsymbol{T_{\text{wall}}-T_{\text{bulk}}} ~^\circ C$ 
& \textbf{Saturation Temp} 
& \textbf{Stefan No.}  \\
\hline
\hline
\multirow{6}{*}{%
\begin{tabular}{c}
OP-250 \\
$T_{\text{bulk}}=41^{\circ}$C \\
$T_{\text{sat}}=49^{\circ}$C
\end{tabular}}
& 81  & 8  & 40 & 0.2000 & 0.3792 \\
& 85  & 12 & 44 & 0.1818 & 0.4171 \\
& 89  & 16 & 48 & 0.1667 & 0.4550 \\
& 93  & 20 & 52 & 0.1538 & 0.4930 \\
& 97  & 24 & 56 & 0.1429 & 0.5309 \\
& 101 & 28 & 60 & 0.1333 & 0.5688 \\
\hline
\end{tabular}
\end{table}


Table \ref{tab:fluid_properties} summarizes the thermophysical properties for the different fluids in the BubbleML dataset and Table \ref{tab:simulation_parameters} the corresponding non-dimensional parameters. 

\input{tables/fluid-props}

\input{tables/nondim-params}


%% file: tables/fluid-props.tex
\begin{table}[htbp]
\centering
\caption{
Thermophysical properties of different fluids at saturation under 1 atm. Sources: \textcolor{blue}{\href{https://www.nist.gov/srd/refprop}{NIST Reference Fluid Thermodynamic and Transport Properties Database}} \cite{lemmon2018nist}, \textcolor{blue}{\href{https://www.opteon.com/en/products/liquid-cooling/2p50}{Opteon-2P50 Datasheet}}.}
\small
\begin{tabular}{l|l|c|c|c|c}
\hline
\textbf{Parameters} & \textbf{Units} & \textbf{Opteon 2P50} & \textbf{FC-72} & \textbf{R515b} & \textbf{LN\textsubscript{2}} \\
\hline
\hline
Saturation Temperature ($T_{sat}$) & °C & 49 & 58 & -19 & -196 \\

Liquid Density ($\rho_l$) & kg·m\textsuperscript{-3} & 1450 & 1575.6 & 1313.7 & 807 \\

Vapor Density ($\rho_v$) & kg·m\textsuperscript{-3} & 8.7 & 13.687 & 5.8361 & 4.51 \\

Liquid Viscosity ($\mu_l$) & N·s·m\textsuperscript{-2} & 6.2$\times$10\textsuperscript{-4} & 4.18$\times$10\textsuperscript{-4} & 3.427$\times$10\textsuperscript{-4} & 1.62$\times$10\textsuperscript{-4} \\

Vapor Viscosity ($\mu_v$) & N·s·m\textsuperscript{-2} & 1.5 $\times$ 10\textsuperscript{-5} & 1.177$\times$10\textsuperscript{-5} & 9.626$\times$10\textsuperscript{-6} & 5.428$\times$10\textsuperscript{-6} \\

Liquid Specific Heat Capacity ($C_{p_l}$) & J·kg\textsuperscript{-1}·K\textsuperscript{-1} & 1090 & 1099.5 & 1263.6 & 2040.5 \\

Vapor Specific Heat Capacity ($C_{p_v}$) & J·kg\textsuperscript{-1}·K\textsuperscript{-1} & 780 & 879.30 & 823.26 & 1122.4 \\

Liquid Thermal Conductivity ($k_l$) & W·m\textsuperscript{-1}·K\textsuperscript{-1} & 7.3$\times$10\textsuperscript{-2} & 6.25$\times$10\textsuperscript{-2} & 8.887$\times$10\textsuperscript{-2} & 0.145 \\

Vapor Thermal Conductivity ($k_v$) & W·m\textsuperscript{-1}·K\textsuperscript{-1} & 1.1$\times$10\textsuperscript{-2} & 1.306$\times$10\textsuperscript{-2} & 1.029$\times$10\textsuperscript{-2} & 7.163$\times$10\textsuperscript{-3} \\

Latent Heat of Vaporization ($h_{lv}$) & J·kg\textsuperscript{-1} & 1.15$\times$10\textsuperscript{5} & 8.4227$\times$10\textsuperscript{4} & 1.9056$\times$10\textsuperscript{5} & 1.9944$\times$10\textsuperscript{5} \\

Surface Tension ($\sigma$) & N·m\textsuperscript{-1} & 1.1$\times$10\textsuperscript{-2} & 8.112$\times$10\textsuperscript{-3} & 1.499$\times$10\textsuperscript{-2} & 8.926$\times$10\textsuperscript{-3} \\

\hline
\end{tabular}
\label{tab:fluid_properties}
\end{table}

%% file: tables/nondim-params.tex
\begin{table}[htbp]
\centering
\caption{Non-dimensional parameters for different fluids at saturation under 1 atm. $\Delta T$ is the difference between the heater temperature and bulk liquid temperature.}
\small
\begin{tabular}{l|c|c|c|c|c}
\hline
\textbf{Simulation Parameter} & \textbf{Formula} & \textbf{OP250} & \textbf{FC-72} & \textbf{R515b} & \textbf{LN\textsubscript{2}} \\
\hline
\hline
Characteristic Length ($l_c$) (mm) & $\sqrt{\frac{\sigma}{(\rho_l - \rho_v)g}}$ & 0.88 & 0.73 & 1.08 & 1.06 \\

Characteristic Velocity ($u_c$)  (m·s\textsuperscript{-1}) & $\sqrt{gl_c}$ & 0.093 & 0.08 & 0.1 & 0.1 \\

Characteristic Time ($t_c$) (ms) & $\frac{l_c}{u_c}$ & 9.5 & 8.6 & 10.5 & 10.4 \\

Density ($\rho'$) & $\frac{\rho_v}{\rho_l}$ & 6.0$\times$10\textsuperscript{-3} & 8.687$\times$10\textsuperscript{-3} & 4.442$\times$10\textsuperscript{-3} & 5.589$\times$10\textsuperscript{-3} \\

Viscosity ($\mu'$) & $\frac{\mu_v}{\mu_l}$ & 2.42$\times$10\textsuperscript{-2} & 2.816$\times$10\textsuperscript{-2} & 2.809$\times$10\textsuperscript{-2} & 3.351$\times$10\textsuperscript{-2} \\

Thermal Conductivity ($k'$) & $\frac{k_v}{k_l}$ & 1.51$\times$10\textsuperscript{-1} & 2.09$\times$10\textsuperscript{-1} & 1.158$\times$10\textsuperscript{-1} & 4.94$\times$10\textsuperscript{-2} \\

Specific Heat ($C_p'$) & $\frac{C_{pv}}{C_{pl}}$ & 7.16$\times$10\textsuperscript{-1} & 7.997$\times$10\textsuperscript{-1} & 6.515$\times$10\textsuperscript{-1} & 5.501$\times$10\textsuperscript{-1} \\

Reynolds Number (Re) & $\frac{\rho_l u_c l_c}{\mu_l}$ & 238 & 231.72 & 426.67 & 542.13 \\

Weber Number (We) & $\frac{\rho_l u_c^2 l_c}{\sigma}$ & 1.0 & 1.0 & 1.0 & 1.0 \\

Prandtl Number (Pr) & $\frac{\mu_l C_{pl}}{k_l}$ & 7.47 & 7.35 & 4.87 & 2.28 \\

Stefan Number (St) & $\frac{C_{pl}}{h_{lv}}\Delta T$ & 0.0095$\Delta T$& 0.013$\Delta T$ & 0.0066$\Delta T$ & 0.0102$\Delta T$ \\
\hline
\end{tabular}
\label{tab:simulation_parameters}
\end{table}

%% file: appendix/addn-results.tex
\section{Ablation: SDF Reinitialization vs Training with Eikonal Loss}
\label{app:ablation-sdf}

We compare (a) test-time SDF reinitialization using the Sussman method with (b) training with an Eikonal loss at varying weights. I.e., $L_{total} = L_{data} + w \times   L_{eik}$, where $w$ is a hyperparameter scaling the loss weight. 
The Eikonal loss is defined as $L_{eik}(\phi_t) = 
 \text{abs}(|\nabla \phi_t| - 1)$ for the SDF $\phi_t$ at every timestep $t$. We sweep weights $w \in \{5, 3, 1, 0.5, 0.1, 0.01, 0\}$.
Table \ref{tab:sdf-reinit-vs-eikonal-loss} reports the mean and maximum errors across all timesteps for two simulations.

\begin{table}[h]
\centering
\caption{Comparison of SDF reinitialization via the Sussman method and training with different Eikonal loss weights.}
\label{tab:sdf-reinit-vs-eikonal-loss}

\begin{tabular}{llcccccccc}
\toprule
& & & & \multicolumn{6}{c}{Eikonal Loss Weight $w$} \\
\cmidrule(lr){5-10}

Case & Metric & Sim. SDF & Sussman Reinit
& $5$ & $3$ & $1$ & $0.5$ & $0.1$ & $0.01$ \\

\midrule

\multirow{2}{6em}[-.1ex]{OP2P50 93 $^\circ$C}
& Mean Err & 0.038 & \textbf{0.049}
& 1.189 & 1.046 & 0.810 & 0.279 & 0.062 & 0.091 \\

& Max Err & 0.08 & \textbf{0.09}
& 2.775 & 2.417 & 1.947 & 0.796 & 0.283 & 0.206 \\

\midrule

\multirow{2}{6em}[-.1ex]{OP2P50 97 $^\circ$C}
& Mean Err & 0.043 & 0.072
& 1.005 & 0.855 & 0.535 & 0.202 & \textbf{0.05} & 0.099 \\

& Max Err & 0.084 & \textbf{0.131}
& 2.480 & 1.994 & 1.349 & 0.454 & 0.168 & 0.228 \\

\bottomrule
\end{tabular}
\end{table}

These results indicate that a \emph{properly weighted} Eikonal loss ($w=0.1$) can achieve mean errors comparable to reinitialization. However, it is very sensitive to the choice of loss weight (errors are catastrophically large for $w \geq 1$). The sensitivity to the Eikonal loss weight is a problem because one has to perform additional training runs simply to approach the results of a cheap SDF reinitialization. In both cases, the max error of reinitialization is lower than any of the models trained with an additional Eikonal loss term indicating better stability. Max error also reflects the worst-case at bubble interfaces and heater surface, which is precisely where SDF accuracy is critical for downstream cooling and energy applications. Incorporating the Sussman method into the training loop is a potential direction for future work.

\section{Model Parameter Counts}
\label{app:parameters}

\input{tables/parameter_counts}

\section{Comparison of Single-step and Rollout Performance}
\label{app:viz-baselines}

\begin{figure}[h]
    \centering
    \includegraphics[width=0.4\linewidth]{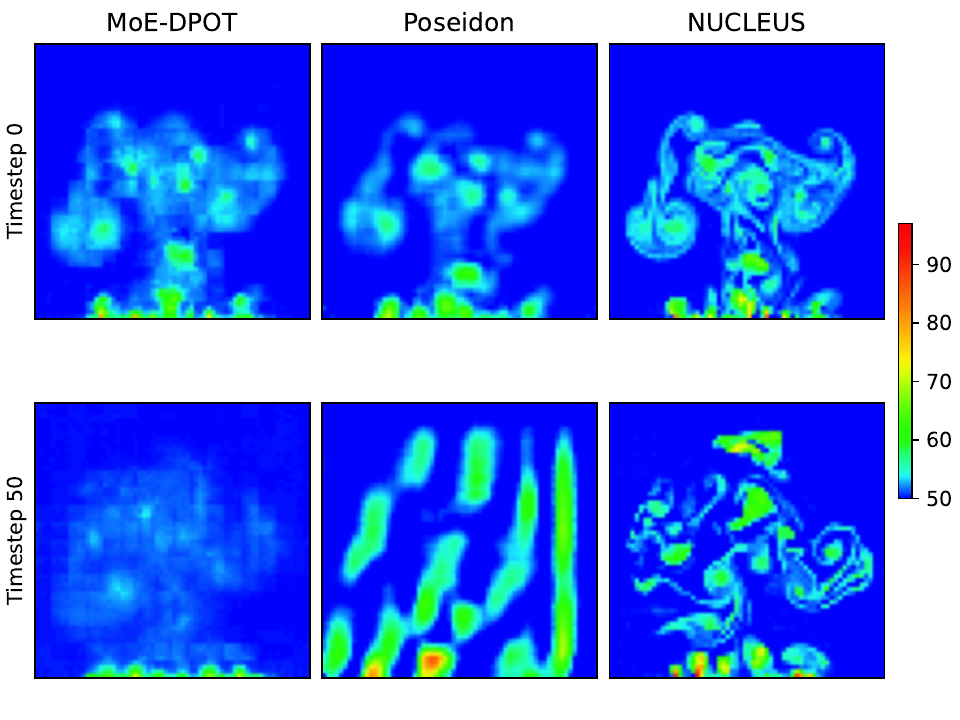}
    \caption{Comparison of MoE-DPOT, Poseidon, and \ours. Both MoE-DPOT and Poseidon exhibit diffused thermal structures and loss of interface sharpness even at single inference step. During autoregressive rollouts, errors compound and performance deteriorates rapidly. In contrast, \ours better preserves coherent thermal and interface structures.}
    \label{fig:temp-comp}
    \vspace{-3em}
\end{figure}

%% file: tables/parameter_counts.tex
\begin{table}[h]
    \centering
    \caption{Parameter counts (millions) for each model. MoE architectures report activated/total parameters. \ours achieves the best overall performance with fewer activated parameters, indicating improved compute efficiency.}

    \begin{tabular}{lr}
    \toprule
        Model   & Parameters \\
    \midrule
        NUCLEUS & 76M / 224M \\
        NAtten + MLP & 115M \\
        Bubbleformer & 115M \\
        MoE-DPOT & 66M / 123M \\
        Poseidon ScOT & 628M \\
    \bottomrule
    \end{tabular}
    \label{tab:placeholder}
\end{table}

%% file: main.bib
@article{sussman1999reinit,
	author = {Sussman, Mark and Fatemi, Emad},
	doi = {10.1137/S1064827596298245},
	eprint = {https://doi.org/10.1137/S1064827596298245},
	journal = {SIAM Journal on Scientific Computing},
	number = {4},
	pages = {1165-1191},
	title = {An Efficient, Interface-Preserving Level Set Redistancing Algorithm and Its Application to Interfacial Incompressible Fluid Flow},
	url = {https://doi.org/10.1137/S1064827596298245},
	volume = {20},
	year = {1999}}

@techreport{congress,
    author={Lawson, Ashley and Offutt, Martin and Ortiz, Natalie and Zhu, Ling},
    title={Data Centers and Their Energy Consumption: Frequently Asked Questions},
    institution={U.S. Congress},
    number={R48646},
    year={2026},
    url={https://www.congress.gov/crs-product/R48646}
}

@book{pope2000turbulentflows, 
    place={Cambridge}, 
    title={Turbulent Flows}, 
    publisher={Cambridge University Press}, 
    author={Pope, Stephen B.}, 
    year={2000}
}

@article{hassan2023bubbleml,
  title={Bubbleml: A multiphase multiphysics dataset and benchmarks for machine learning},
  author={Hassan, Sheikh Md Shakeel and Feeney, Arthur and Dhruv, Akash and Kim, Jihoon and Suh, Youngjoon and Ryu, Jaiyoung and Won, Yoonjin and Chandramowlishwaran, Aparna},
  journal={Advances in Neural Information Processing Systems},
  volume={36},
  pages={418--449},
  year={2023}
}

@article{hassan2025bubbleformer,
  title={Bubbleformer: Forecasting Boiling with Transformers},
  author={Hassan, Sheikh Md Shakeel and Zou, Xianwei and Dhruv, Akash and Chandramowlishwaran, Aparna},
  journal={Advances in Neural Information Processing Systems},
  volume={38},
  year={2026}
}

@inproceedings{
wang2025mixtureofexperts,
title={Mixture-of-Experts Operator Transformer for Large-Scale {PDE} Pre-Training},
author={Hong Wang and Haiyang Xin and Jie Wang and Xuanze Yang and Fei Zha and huanshuo dong and Yan Jiang},
booktitle={The Thirty-ninth Annual Conference on Neural Information Processing Systems},
year={2025},
url={https://openreview.net/forum?id=PNgG4H3q9D}
}

@article{lemmon2018nist,
  title={NIST standard reference database 23: reference fluid thermodynamic and transport properties-REFPROP, Version 10.0, National Institute of Standards and Technology},
  author={Lemmon, Eric W and Bell, Ian H and Huber, ML and McLinden, MO},
  journal={Standard Reference Data Program, Gaithersburg},
  pages={45--46},
  year={2018}
}

@inproceedings{hassani2023neighborhood,
  title={Neighborhood attention transformer},
  author={Hassani, Ali and Walton, Steven and Li, Jiachen and Li, Shen and Shi, Humphrey},
  booktitle={Proceedings of the IEEE/CVF conference on computer vision and pattern recognition},
  pages={6185--6194},
  year={2023}
}

@article{kossaifi2026demystifying,
  title={Demystifying Data-Driven Probabilistic Medium-Range Weather Forecasting},
  author={Kossaifi, Jean and Kovachki, Nikola and Mardani, Morteza and Leibovici, Daniel and Ravuri, Suman and Shokar, Ira and Calvello, Edoardo and Abbas, Mohammad Shoaib and Harrington, Peter and Subramaniam, Ashay and others},
  journal={arXiv preprint arXiv:2601.18111},
  year={2026}
}

@inproceedings{
shazeer2017,
title={ Outrageously Large Neural Networks: The Sparsely-Gated Mixture-of-Experts Layer},
author={Noam Shazeer and *Azalia Mirhoseini and *Krzysztof Maziarz and Andy Davis and Quoc Le and Geoffrey Hinton and Jeff Dean},
booktitle={International Conference on Learning Representations},
year={2017},
url={https://openreview.net/forum?id=B1ckMDqlg}
}

@book{osher2003level,
    title={Level Set Methods and Dynamic Implicit Surfaces},
    author={Osher, Stanley and Fedkiw, Ronald},
    year={2003},
    publisher={Springer-Verlang New York Inc.}
}

@article{Dubey_2022,
   title={Flash-X: A multiphysics simulation software instrument},
   volume={19},
   ISSN={2352-7110},
   url={http://dx.doi.org/10.1016/j.softx.2022.101168},
   DOI={10.1016/j.softx.2022.101168},
   journal={SoftwareX},
   publisher={Elsevier BV},
   author={Dubey, Anshu and Weide, Klaus and O’Neal, Jared and Dhruv, Akash and Couch, Sean and Harris, J. Austin and Klosterman, Tom and Jain, Rajeev and Rudi, Johann and Messer, Bronson and Pajkos, Michael and Carlson, Jared and Chu, Ran and Wahib, Mohamed and Chawdhary, Saurabh and Ricker, Paul M. and Lee, Dongwook and Antypas, Katie and Riley, Katherine M. and Daley, Christopher and Ganapathy, Murali and Timmes, Francis X. and Townsley, Dean M. and Vanella, Marcos and Bachan, John and Rich, Paul M. and Kumar, Shravan and Endeve, Eirik and Hix, W. Raphael and Mezzacappa, Anthony and Papatheodore, Thomas},
   year={2022},
   month=jul, pages={101168} }

@misc{li2021fourierneuraloperatorparametric,
      title={Fourier Neural Operator for Parametric Partial Differential Equations}, 
      author={Zongyi Li and Nikola Kovachki and Kamyar Azizzadenesheli and Burigede Liu and Kaushik Bhattacharya and Andrew Stuart and Anima Anandkumar},
      year={2021},
      eprint={2010.08895},
      archivePrefix={arXiv},
      primaryClass={cs.LG},
      url={https://arxiv.org/abs/2010.08895}, 
}

@misc{fedus2022switchtransformersscalingtrillion,
      title={Switch Transformers: Scaling to Trillion Parameter Models with Simple and Efficient Sparsity}, 
      author={William Fedus and Barret Zoph and Noam Shazeer},
      year={2022},
      eprint={2101.03961},
      archivePrefix={arXiv},
      primaryClass={cs.LG},
      url={https://arxiv.org/abs/2101.03961}, 
}

@article{khodakarami2025mitigating,
  title={Mitigating spectral bias in neural operators via high-frequency scaling for physical systems},
  author={Khodakarami, Siavash and Oommen, Vivek and Bora, Aniruddha and Karniadakis, George Em},
  journal={arXiv preprint arXiv:2503.13695},
  year={2025}
}

@inproceedings{
wang2022antioversmoothing,
title={Anti-Oversmoothing in Deep Vision Transformers via the Fourier Domain Analysis: From Theory to Practice},
author={Peihao Wang and Wenqing Zheng and Tianlong Chen and Zhangyang Wang},
booktitle={International Conference on Learning Representations},
year={2022},
url={https://openreview.net/forum?id=O476oWmiNNp}
}

@article{herde2024poseidon,
  title={Poseidon: Efficient foundation models for pdes},
  author={Herde, Maximilian and Raonic, Bogdan and Rohner, Tobias and K{\"a}ppeli, Roger and Molinaro, Roberto and de B{\'e}zenac, Emmanuel and Mishra, Siddhartha},
  journal={Advances in Neural Information Processing Systems},
  volume={37},
  pages={72525--72624},
  year={2024}
}

@inproceedings{
tran2023factorized,
title={Factorized Fourier Neural Operators},
author={Alasdair Tran and Alexander Mathews and Lexing Xie and Cheng Soon Ong},
booktitle={The Eleventh International Conference on Learning Representations },
year={2023},
url={https://openreview.net/forum?id=tmIiMPl4IPa}
}

@inproceedings{
wu2024transolver,
title={Transolver: A Fast Transformer Solver for {PDE}s on General Geometries},
author={Haixu Wu and Huakun Luo and Haowen Wang and Jianmin Wang and Mingsheng Long},
booktitle={Forty-first International Conference on Machine Learning},
year={2024}
}

@article{brunton2020machine,
  title={Machine learning for fluid mechanics},
  author={Brunton, Steven L and Noack, Bernd R and Koumoutsakos, Petros},
  journal={Annual review of fluid mechanics},
  volume={52},
  number={1},
  pages={477--508},
  year={2020},
  publisher={Annual Reviews}
}

@article{azizzadenesheli2024neural,
  title={Neural operators for accelerating scientific simulations and design},
  author={Azizzadenesheli, Kamyar and Kovachki, Nikola and Li, Zongyi and Liu-Schiaffini, Miguel and Kossaifi, Jean and Anandkumar, Anima},
  journal={Nature Reviews Physics},
  volume={6},
  number={5},
  pages={320--328},
  year={2024},
  publisher={Nature Publishing Group UK London}
}

@article{eyring2024pushing,
  title={Pushing the frontiers in climate modelling and analysis with machine learning},
  author={Eyring, Veronika and Collins, William D and Gentine, Pierre and Barnes, Elizabeth A and Barreiro, Marcelo and Beucler, Tom and Bocquet, Marc and Bretherton, Christopher S and Christensen, Hannah M and Dagon, Katherine and others},
  journal={Nature Climate Change},
  volume={14},
  number={9},
  pages={916--928},
  year={2024},
  publisher={Nature Publishing Group UK London}
}

@article{nukiyama1966maximum,
  title={The maximum and minimum values of the heat Q transmitted from metal to boiling water under atmospheric pressure},
  author={Nukiyama, Shiro},
  journal={International Journal of Heat and Mass Transfer},
  volume={9},
  number={12},
  pages={1419--1433},
  year={1966},
  publisher={Elsevier}
}

@article{rohsenow1952method,
  title={A method of correlating heat-transfer data for surface boiling of liquids},
  author={Rohsenow, Warren M},
  journal={Transactions of the American Society of Mechanical Engineers},
  volume={74},
  number={6},
  pages={969--975},
  year={1952},
  publisher={American Society of Mechanical Engineers}
}

@book{zuber1959hydrodynamic,
  title={Hydrodynamic aspects of boiling heat transfer},
  author={Zuber, Novak},
  number={4439},
  year={1959},
  publisher={United States Atomic Energy Commission, Technical Information Service}
}

@phdthesis{han1962mechanism,
  title={The mechanism of heat transfer in nucleate pool boiling},
  author={Han, Chi-Yeh},
  year={1962},
  school={Massachusetts Institute of Technology}
}

@article{dhir2013numerical,
  title={Numerical simulation of pool boiling: a review},
  author={Dhir, Vijay K and Warrier, Gopinath R and Aktinol, Eduardo},
  journal={Journal of Heat Transfer},
  volume={135},
  number={6},
  pages={061502},
  year={2013},
  publisher={American Society of Mechanical Engineers}
}

@article{sato2018pool,
  title={Pool boiling simulation using an interface tracking method: From nucleate boiling to film boiling regime through critical heat flux},
  author={Sato, Yohei and Niceno, Bojan},
  journal={International Journal of Heat and Mass Transfer},
  volume={125},
  pages={876--890},
  year={2018},
  publisher={Elsevier}
}

@article{dhruv2019formulation,
  title={A formulation for high-fidelity simulations of pool boiling in low gravity},
  author={Dhruv, Akash and Balaras, Elias and Riaz, Amir and Kim, Jungho},
  journal={International Journal of Multiphase Flow},
  volume={120},
  pages={103099},
  year={2019},
  publisher={Elsevier}
}

@misc{lippe2023pderefinerachievingaccuratelong,
      title={PDE-Refiner: Achieving Accurate Long Rollouts with Neural PDE Solvers}, 
      author={Phillip Lippe and Bastiaan S. Veeling and Paris Perdikaris and Richard E. Turner and Johannes Brandstetter},
      year={2023},
      eprint={2308.05732},
      archivePrefix={arXiv},
      primaryClass={cs.LG},
      url={https://arxiv.org/abs/2308.05732}, 
}

@article{krajewski2024scaling,
  title={Scaling laws for fine-grained mixture of experts},
  author={Krajewski, Jakub and Ludziejewski, Jan and Adamczewski, Kamil and Pi{\'o}ro, Maciej and Krutul, Micha{\l} and Antoniak, Szymon and Ciebiera, Kamil and Kr{\'o}l, Krystian and Odrzyg{\'o}{\'z}d{\'z}, Tomasz and Sankowski, Piotr and others},
  journal={arXiv preprint arXiv:2402.07871},
  year={2024}
}

@article{riquelme2021scaling,
  title={Scaling vision with sparse mixture of experts},
  author={Riquelme, Carlos and Puigcerver, Joan and Mustafa, Basil and Neumann, Maxim and Jenatton, Rodolphe and Susano Pinto, Andr{\'e} and Keysers, Daniel and Houlsby, Neil},
  journal={Advances in Neural Information Processing Systems},
  volume={34},
  pages={8583--8595},
  year={2021}
}

@article{obando2024mixtures,
  title={Mixtures of experts unlock parameter scaling for deep rl},
  author={Obando-Ceron, Johan and Sokar, Ghada and Willi, Timon and Lyle, Clare and Farebrother, Jesse and Foerster, Jakob and Dziugaite, Gintare Karolina and Precup, Doina and Castro, Pablo Samuel},
  journal={arXiv preprint arXiv:2402.08609},
  year={2024}
}

@inproceedings{
chalapathi2024scaling,
title={Scaling physics-informed hard constraints with mixture-of-experts},
author={Nithin Chalapathi and Yiheng Du and Aditi S. Krishnapriyan},
booktitle={The Twelfth International Conference on Learning Representations},
year={2024},
url={https://openreview.net/forum?id=u3dX2CEIZb}
}

@article{dai2024deepseekmoe,
  title={Deepseekmoe: Towards ultimate expert specialization in mixture-of-experts language models},
  author={Dai, Damai and Deng, Chengqi and Zhao, Chenggang and Xu, RX and Gao, Huazuo and Chen, Deli and Li, Jiashi and Zeng, Wangding and Yu, Xingkai and Wu, Yu and others},
  journal={arXiv preprint arXiv:2401.06066},
  year={2024}
}

@article{bucci2016mechanistic,
  title={A mechanistic IR calibration technique for boiling heat transfer investigations},
  author={Bucci, Matteo and Richenderfer, Andrew and Su, Guan-Yu and McKrell, Thomas and Buongiorno, Jacopo},
  journal={International Journal of Multiphase Flow},
  volume={83},
  pages={115--127},
  year={2016},
  publisher={Elsevier}
}

@article{liu2020uncertainty,
  title={Uncertainty analysis of PIV measurements in bubbly flows considering sampling and bubble effects with ray optics modeling},
  author={Liu, Yang and Wang, Chengqi and Qian, Yalan and Sun, Xiaodong},
  journal={Nuclear Engineering and Design},
  volume={364},
  pages={110677},
  year={2020},
  publisher={Elsevier}
}

@book{faghri2019fundamentals,
  title={Fundamentals of Multiphase Heat Transfer and Flow},
  author={Faghri, Amir and Zhang, Yuwen},
  year={2019},
  publisher={Springer Nature}
}

@article{dirker2019thermal,
  title={Thermal energy processes in direct steam generation solar systems: Boiling, condensation and energy storage--A review},
  author={Dirker, Jaco and Juggurnath, Diksha and Kaya, Alihan and Osowade, Emmanuel A and Simpson, Michael and Lecompte, Steven and Noori Rahim Abadi, Seyyed Mohammad Ali and Voulgaropoulos, Victor and Adelaja, Adekunle O and Dauhoo, M Zaid and others},
  journal={Frontiers in Energy Research},
  volume={6},
  pages={147},
  year={2019},
  publisher={Frontiers Media SA}
}

@article{azarifar2024liquid,
  title={Liquid cooling of data centers: A necessity facing challenges},
  author={Azarifar, Mohammad and Arik, Mehmet and Chang, Je-Young},
  journal={Applied Thermal Engineering},
  volume={247},
  pages={123112},
  year={2024},
  publisher={Elsevier}
}

@article{wang2025cross,
  title={Cross-Field Interface-Aware Neural Operators for Multiphase Flow Simulation},
  author={Wang, Zhenzhong and Zhang, Xin and Liao, Jun and Jiang, Min},
  journal={arXiv preprint arXiv:2511.08625},
  year={2025}
}

@article{shadkhah2025mpfbench,
  title={MPFBench: A Large Scale Dataset for SciML of Multi-Phase-Flows: Droplet and Bubble Dynamics},
  author={Shadkhah, Mehdi and Tali, Ronak and Rabeh, Ali and Yang, Cheng-Hau and Herron, Ethan and Upadhyaya, Abhisek and Krishnamurthy, Adarsh and Hegde, Chinmay and Balu, Aditya and Ganapathysubramanian, Baskar},
  journal={arXiv preprint arXiv:2502.07080},
  year={2025}
}

@article{dosovitskiy2020image,
  title={An image is worth 16x16 words: Transformers for image recognition at scale},
  author={Dosovitskiy, Alexey},
  journal={arXiv preprint arXiv:2010.11929},
  year={2020}
}

@inproceedings{perez2018film,
  title={Film: Visual reasoning with a general conditioning layer},
  author={Perez, Ethan and Strub, Florian and De Vries, Harm and Dumoulin, Vincent and Courville, Aaron},
  booktitle={Proceedings of the AAAI conference on artificial intelligence},
  volume={32},
  number={1},
  year={2018}
}

@inproceedings{
poels2024accelerating,
title={Accelerating Simulation of Two-Phase Flows with Neural {PDE} Surrogates},
author={Yoeri Poels and Koen Minartz and Harshit Bansal and Vlado Menkovski},
booktitle={ICML 2024 AI for Science Workshop},
year={2024},
url={https://openreview.net/forum?id=yIqszw9RUc}
}
